\documentclass[10pt,twocolumn,letterpaper]{article}

\usepackage[pagenumbers]{cvpr} % To force page numbers, e.g. for an arXiv version

\usepackage{graphicx}
\usepackage{amsmath}
\usepackage{amssymb}
\usepackage{booktabs}
\usepackage{multirow}
\usepackage{comment}
\usepackage{xcolor}         % colors
\usepackage[accsupp]{axessibility}  % Improves PDF readability for those with disabilities.
\usepackage[linesnumbered,lined,vlined,ruled,commentsnumbered]{algorithm2e}

\newcommand{\myparagraphsupp}[1]{\vspace{0.1em}\noindent{\textcolor{red}{#1}}}
\newcommand{\mycaptionsupp}[1]{{\textcolor{red}{#1}}}

\newcommand{\beginsupp}{%
        \setcounter{table}{0}
        \renewcommand{\thetable}{S\arabic{table}}%
        \setcounter{figure}{0}
        \renewcommand{\thefigure}{S\arabic{figure}}%
        \setcounter{equation}{0}
        \renewcommand{\theequation}{S\arabic{equation}}%
        \setcounter{section}{0}
        \renewcommand\thesection{\Alph{section}}
     }

\usepackage[pagebackref,breaklinks,colorlinks]{hyperref}

\usepackage[capitalize]{cleveref}
\crefname{section}{Sec.}{Secs.}
\Crefname{section}{Section}{Sections}
\Crefname{table}{Table}{Tables}
\crefname{table}{Tab.}{Tabs.}

\def\cvprPaperID{9895} % *** Enter the CVPR Paper ID here
\def\confName{CVPR}
\def\confYear{2023}

\begin{document}

\title{Freestyle Layout-to-Image Synthesis}

\author{Han Xue$^{1,2}$ \quad Zhiwu Huang$^{2,3}$ \quad Qianru Sun$^{2}$ \quad Li Song$^{1,4}$\footnotemark[1] \quad Wenjun Zhang$^{1}$\\
$^{1}$School of Electronic Information and Electrical Engineering, Shanghai Jiao Tong University\\
$^{2}$Singapore Management University \quad $^{3}$University of Southampton\\
$^{4}$MoE Key Lab of Artificial Intelligence, AI Institute, Shanghai Jiao Tong University\\
\tt\small \{xue\_han, song\_li, zhangwenjun\}@sjtu.edu.cn \ \tt\small qianrusun@smu.edu.sg \ \tt\small zhiwu.huang@soton.ac.uk
}

\twocolumn[{
\maketitle
\begin{center}
    \captionsetup{type=figure}
    \includegraphics[width=0.94\linewidth]{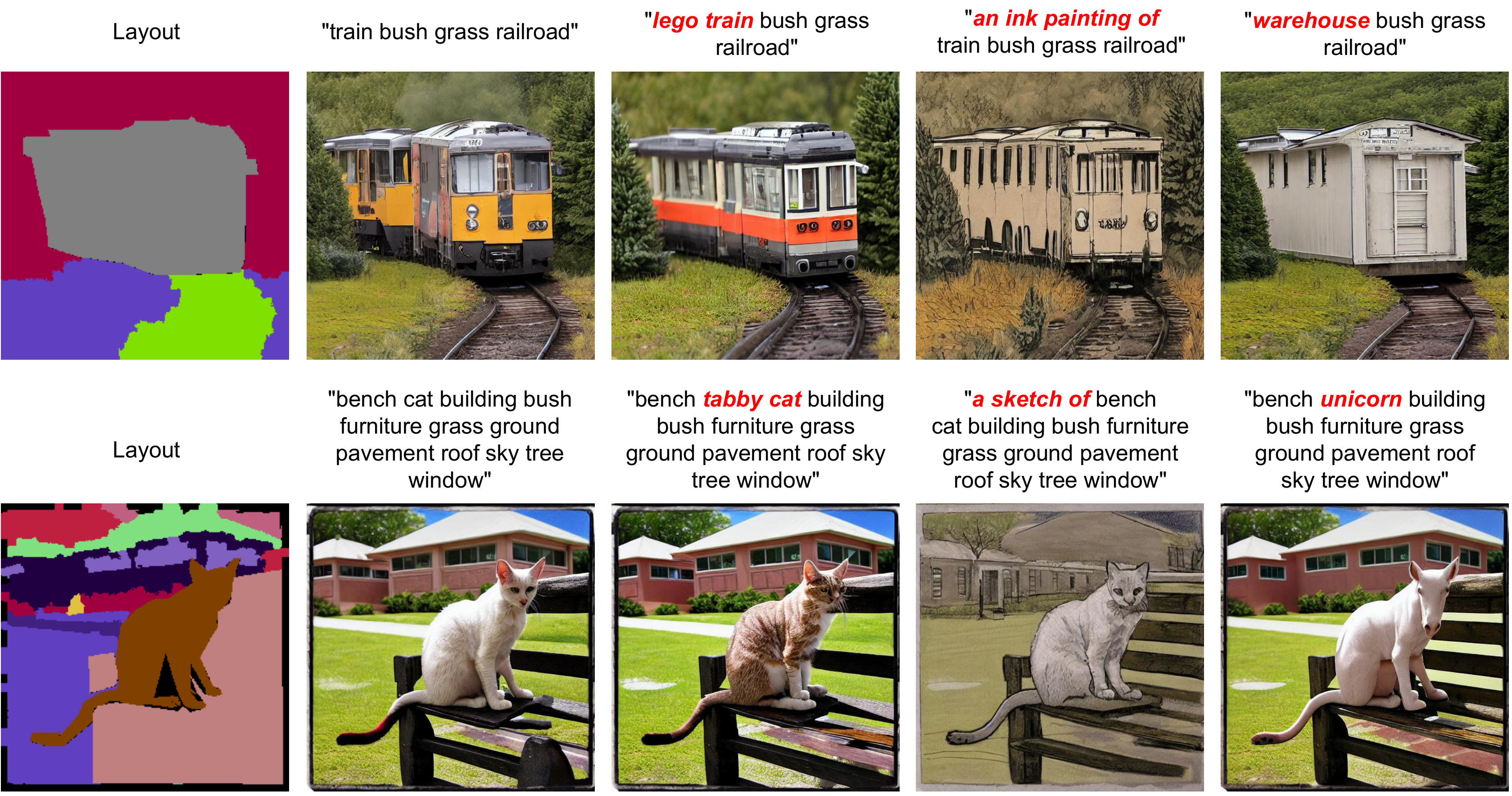}
    \vspace{-0.3cm}
    \captionof{figure}{Freestyle Layout-to-Image Synthesis (FLIS) results generated by using our model. 
    Each has two kinds of inputs: a layout of semantic masks (on the 1st column), and a text (on the top of each result).
    For each layout, we show three example results with edited texts (3rd-5th columns).
    They validate that our model is able to introduce new attributes (3rd column), styles (4th column), and objects (5th column), which are all unseen during training, in the synthesized images. The generated hornless unicorn is due to the layout constraint.
    }
    \label{fig:teaser}
\end{center}
}]
\renewcommand{\thefootnote}{\fnsymbol{footnote}} 
\footnotetext[1]{Corresponding author. Part of this work was done when Han Xue served as a visiting Ph.D. student at Singapore Management University.}
% \maketitle

%%%%%%%%% ABSTRACT
\begin{abstract}

Typical layout-to-image synthesis (LIS) models generate images for a closed set of semantic classes, e.g., $182$ common objects in COCO-Stuff.
In this work, we explore the freestyle capability of the model, i.e., how far can it generate unseen semantics (e.g., classes, attributes, and styles) onto a given layout, and call the task Freestyle LIS (FLIS).
Thanks to the development of large-scale pre-trained language-image models, a number of discriminative models (e.g., image classification and object detection) trained on limited base classes are empowered with the ability of unseen class prediction.
Inspired by this, we opt to leverage large-scale pre-trained text-to-image diffusion models to achieve the generation of unseen semantics.
The key challenge of FLIS is how to enable the diffusion model to synthesize images from a specific layout which very likely violates its pre-learned knowledge, e.g., the model never sees ``a unicorn sitting on a bench'' during its pre-training.
To this end, we introduce a new module called Rectified Cross-Attention (RCA) that can be conveniently plugged in the diffusion model to integrate semantic masks.
This ``plug-in'' is applied in each cross-attention layer of the model to rectify the attention maps between image and text tokens.
The key idea of RCA is to enforce each text token to act on the pixels in a specified region, allowing us to freely put a wide variety of semantics from pre-trained knowledge (which is general) onto the given layout (which is specific).
Extensive experiments show that the proposed diffusion network produces realistic and freestyle layout-to-image generation results with diverse text inputs, which has a high potential to spawn a bunch of interesting applications.
Code is available at \href{https://github.com/essunny310/FreestyleNet}{https://github.com/essunny310/FreestyleNet}.

\end{abstract}

%%%%%%%%% BODY TEXT
\section{Introduction}\label{sec:intro}

Layout-to-image synthesis (LIS) is one of the prevailing topics in the research of conditional image generation.
It aims to generate complex scenes where a user requires fine controls over the objects appearing in a scene.
There are different types of layouts including bboxes+classes~\cite{hinz2019generating}, semantic masks+classes~\cite{park2019semantic,oasis}, key points+attributes~\cite{reed2016learning}, skeletons~\cite{lv2021learning}, to name a few.
In this work, we focus on using semantic masks to control object shapes and positions, and using texts to control classes, attributes, and styles.
We aim to offer the user the most fine-grained controls on output images.
For example, in Figure~\ref{fig:teaser} (2nd column), the train and the cat are generated onto the right locations defined by the input semantic layouts, and they should also be aligned well with their surroundings.

However, most existing LIS models~\cite{wang2018high,park2019semantic,liu2019learning,oasis,wang2021image} are trained on a specific dataset from scratch. Therefore, the generated images are limited to a closed set of semantic classes, \eg, 182 objects on COCO-Stuff~\cite{caesar2018coco}. While in this paper, we explore the \textit{freestyle} capability of the LIS models for the generation of unseen classes (including their attributes and styles) onto a given layout.
For instance, in Figure~\ref{fig:teaser} (5th column), our model stretches and squeezes a ``warehouse'' into a mask of ``train'', and the ``warehouse'' is never seen in the training dataset of LIS. In addition, 
it is also able to introduce novel attributes and styles into the synthesized images (in the 3rd and 4th columns of Figure~\ref{fig:teaser}).
The \emph{freestyle} property aims at breaking the in-distribution limit of LIS, and it has a high potential to enlarge the applicability of LIS results to out-of-distribution domains such as data augmentation in open-set or long-tailed semantic segmentation tasks.

To overcome the in-distribution limit, large-scale pre-trained language-image models (\eg, CLIP~\cite{radford2021learning}) 
have shown the ability of learning general knowledge on an enormous amount of image-text pairs. The learned knowledge enables a multitude of discriminative methods trained with limited base classes to predict unseen classes.
For instance, the trained models in~\cite{radford2021learning,zhou2022learning,wang2022sprompt} based on pre-trained CLIP are able to do zero-shot or few-shot classification, which is achieved by finding the most relevant text prompt (\eg, \emph{a photo of a cat}) given an image.
However, it is non-trivial to apply similar ideas to generative models. In classification, simple texts (\eg, \emph{a photo of a cat}) can be conveniently mapped to labels (\eg, \emph{cat}). It is, however, not intuitive to achieve this mapping from compositional texts (\eg, \emph{a train is running on the railroad with grass on both sides}) to synthesized images.
Thanks to the development of pre-training in image generation tasks, we opt to leverage large-scale pre-trained language-image models to the downstream generation tasks.

To this end, we propose to leverage the large-scale pre-trained text-to-image diffusion model to achieve the LIS of unseen semantics. These diffusion models are powerful in generating images of high quality that are in line with the input texts. This can provide us with a generative prior with a wide range of semantics. However, it is challenging to enable the diffusion model to synthesize images from a specific layout which very likely violates its pre-trained knowledge, \eg, the model never sees ``a unicorn sitting on a bench'' during its pre-training. Textual semantics should be correctly arranged in space without impairing their expressiveness.
A seminal attempt, PITI~\cite{wang2022pretraining}, learns an encoder to map the layout to the textual
space of a pre-trained text-to-image model. Although it demonstrates improved quality of the generated images, PITI has two main drawbacks. First, it abandons the text encoder of the pre-trained model, losing the ability to freely control the generation results using different texts. Second, its introduced layout encoder implicitly matches the space of the pre-trained text encoder and hence incurs inferior spatial alignment with input layouts.

In this paper, we propose a simple yet effective framework for the suggested freestyle layout-to-image synthesis (FLIS) task. We introduce a new module called Rectified Cross-Attention (RCA) and plug it into a pre-trained text-to-image diffusion model. This “plug-in” is applied in each cross-attention layer of the diffusion model to integrate the input layout into the generation process. 
In particular, to ensure the semantics (described by text) appear in a region (specified by layout), we find that image tokens in the region should primarily aggregate information from the corresponding text tokens. In light of this, the key idea of RCA is to utilize the layout to rectify the attention maps computed between image and text tokens, allowing us to put desired semantics from the text onto a layout.
In addition, RCA does not introduce any additional parameters into the pre-trained model, making our framework an effective solution to freestyle layout-to-image synthesis.

As shown in Figure~\ref{fig:teaser}, the proposed diffusion network (FreestyleNet) allows the freestyle synthesis of high-fidelity images with novel semantics, including but not limited to: synthesizing unseen objects, binding new attributes to an object, and rendering the image with various styles. Our main contributions can be summarized as follows:
\begin{itemize}
\item We propose a novel LIS task called FLIS. In this task, we exploit large-scale pre-trained text-to-image diffusion models with layout and text as conditions.

\item We introduce a parameter-free RCA module to plug in the diffusion model, allowing us to generate images from a specified layout effectively while taking full advantage of the model's generative priors.

\item Extensive experiments demonstrate the capability of our approach to translate a layout into high-fidelity images with a wide range of novel semantics, which is beyond the reach of existing LIS models.
\end{itemize}

%-------------------------------------------------------------------------
\section{Related Work}\label{sec:related} 

Deep generative models like Generative Adversarial Networks (GANs)~\cite{DBLP:conf/nips/GoodfellowPMXWOCB14,brock2018large,karras2019style} and diffusion models~\cite{sohl2015deep,ho2020denoising,song2020score,song2020improved,song2020denoising,dhariwal2021diffusion} have achieved impressive results for image synthesis. Compared to unconditional synthesis which yields an image from a random noise, conditional image synthesis models~\cite{fan2022frido,zhang2021ufc} allow more controllability over the generation. Among them, semantic image synthesis and text-based image synthesis are the most relevant to this paper.

\noindent\textbf{Semantic Image Synthesis.} This task aims to generate images based on the given semantic layouts~\cite{chen2017photographic,isola2017image,wang2018high,qi2018semi,park2019semantic,liu2019learning,sushko2020you,zhu2020sean,tang2020dual,zhu2020semantically,wang2021image,tan2021efficient,tan2022semantic,shi2022retrieval,lv2022semantic,wang2022semantic,zhu2022label,alaniz2022semantic}. A pioneer work, Pix2Pix~\cite{isola2017image}, utilizes an encoder-decoder generator and a PatchGAN discriminator to perform semantic image synthesis. Pix2PixHD~\cite{wang2018high} enhances Pix2Pix with coarse-to-fine and multi-scale network architectures for high-resolution image synthesis. SPADE~\cite{park2019semantic} suggests modulating the activations in normalization layers using the affine parameters predicted from the input semantic maps. Instead of modifying the features, CC-FPSE~\cite{liu2019learning} and SC-GAN~\cite{wang2021image} learn to produce convolutional kernels and semantic vectors from the semantic maps to condition the generation. OASIS~\cite{sushko2020you} designs a segmentation-based discriminator to synthesize images with a better alignment to the input label maps. Despite their impressive performance, these methods are limited in generating fully controlled images with diverse semantics. While in this paper, we propose a freestyle layout-to-image synthesis framework that is armed with much more controllability, enabling 1) the generation of semantics beyond the pre-defined semantic categories in the training dataset, and 2) the separate modulation of each class in the layout with text. A recent work, PITI~\cite{wang2022pretraining}, leverages a pre-trained diffusion model to facilitate image-to-image translation, achieving impressive results.
However, its generation is restricted by single-modality condition such as mask.

\noindent\textbf{Text-based Image Synthesis.} Generating images from texts is another popular task of image synthesis. Significant progress has been made by exploiting progressive generation~\cite{zhang2017stackgan}, attention mechanisms~\cite{xu2018attngan}, and cross-modal contrastive approaches~\cite{zhang2021cross}. Recent text-to-image synthesis methods such as GLIDE~\cite{nichol2021glide}, DALL-E2~\cite{ramesh2022hierarchical}, Imagen~\cite{saharia2022photorealistic}, Parti~\cite{yu2022scaling}, CogView2~\cite{ding2022cogview2}, and Stable Diffusion~\cite{rombach2022high} are built upon either auto-regressive or diffusion models. Trained on large-scale image-text pairs, these methods demonstrate an astonishing ability to generate high-fidelity images from texts. Nevertheless, with only text as input, they fall short of fine-grained control over the synthesized results (\eg, specifying the layout).
To advance this, several approaches~\cite{reed2016learning,hong2018inferring,hinz2019generating,liang2022layout} attempt to introduce the location of objects and generate images in a hierarchical manner, but the controls are limited to ambiguous locations (indicated by bboxes). In contrast, our method manages to generate semantics with fine-grained (pixel-level) control. PoE-GAN~\cite{huang2022multimodal} and Make-A-Scene~\cite{gafni2022make} accept both text and layouts, but they support only in-distribution generation. In addition, PoE-GAN~\cite{huang2022multimodal} merely performs global text-based manipulations. Our proposed method supports text-guided and out-of-distribution LIS.

In addition, some recent works succeed in applying the pre-trained text-to-image diffusion models to various applications including text-guided image editing~\cite{hertz2022prompt,kawar2022imagic}, image inpainting~\cite{lugmayr2022repaint}, and personalized generation~\cite{gal2022image,ruiz2022dreambooth}, but none of them can fulfill our goal. In this paper, we excavate the general knowledge learned in a pre-trained text-to-image diffusion model to enable the layout-to-image synthesis to be performed in a freestyle manner.
\section{Problem Definition}\label{sec:problem_definition}

The objective of general layout-to-image synthesis (LIS) using masks is to learn a mapping function $f(l)=x$ 
that generates a realistic image $x$ from the input mask $l$.
This mapping function is expected to generate semantics onto the given mask,
and is usually trained in a specific dataset.
The generated semantics are thus highly constrained by the dataset.
To achieve the generation of freestyle semantics,
we define a new task called freestyle layout-to-image synthesis (FLIS), and introduce a feasible solution for it.
Specifically, we introduce to condition the model with not only masks but also texts. Masks control the layout of semantics, while texts elaborate what specific semantics to ``put onto'' the layout. 
Therefore, FLIS mapping function $f$ can be formulated as $f(l,y)=x$, where $l$ and $y$ are the input masks and texts respectively.

An intuitive approach to initializing the input text $y$ is to use the labels of semantic masks in $l$, \eg, ``train bush grass railroad'' in the first example of Figure~\ref{fig:teaser}.
In any specific image generation dataset, 
the vocabulary of these labels is fixed, \eg, 182 objects on COCO-Stuff~\cite{caesar2018coco}.
To go beyond this, we opt to adopt the large-scale diffusion models such as Stable Diffusion \cite{rombach2022high} that have seen far more labels during their pre-training on billions of open data (\eg, LAION-5B \cite{schuhmann2022laion}).
We expect to leverage the general knowledge of such models to enable the generation of open-set semantics in the task of FLIS.

The main challenge is how to synthesize images from a specific layout which very likely violates the pre-trained knowledge of such models, \eg, the model has seen the images of ``unicorn'' but never seen ``the unicorn sitting on a bench'' (with the exact relative location of ``unicorn'' and ``bench" being specified) during its pre-training.
To solve the issue, we propose a framework with Rectified Cross-Attention (RCA) to explicitly ``inject'' a specific layout (a sequence of masks) into the generation process of pre-trained diffusion models.
The key idea is to ensure that the cross-attention of each text token is only performed in the region defined by its corresponding mask. 
The proposed framework will be elaborated in the next section.
\section{Proposed Method}\label{sec:method}
\begin{figure}[t]
   \centering
   \includegraphics[width=1\linewidth]{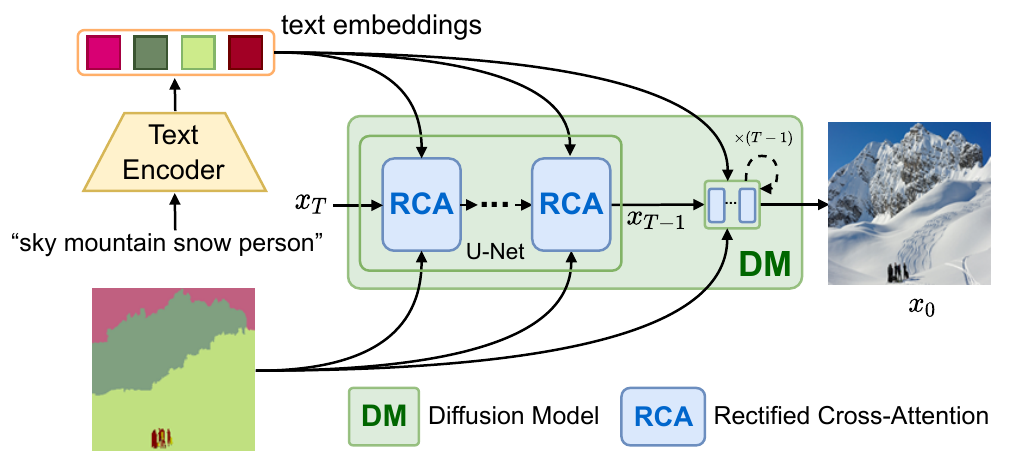}
   %\vspace{-0.1cm}
  \caption{An overview of the proposed FreestyleNet. By applying Rectified Cross-Attention (RCA) in the pre-trained text-to-image diffusion model, Stable Diffusion~\cite{rombach2022high}, our method allows for freestyle layout-to-image synthesis, where a layout can be translated into realistic and creative images with a wide range of semantics specified by the input texts. 
  Note that the base text shown here is simply derived from the semantic layout in order to train our network on image-layout pairs. During inference, we can freely edit the text to generate diverse images.
  Here the text encoder and the diffusion model are adopted from Stable Diffusion and we omit the autoencoder of Stable Diffusion for simplicity.
  }
   \label{fig:framework}
%   \vspace{-0.4cm}
\end{figure} 

\begin{figure*}[t]
   \centering
   \includegraphics[width=0.96\linewidth]{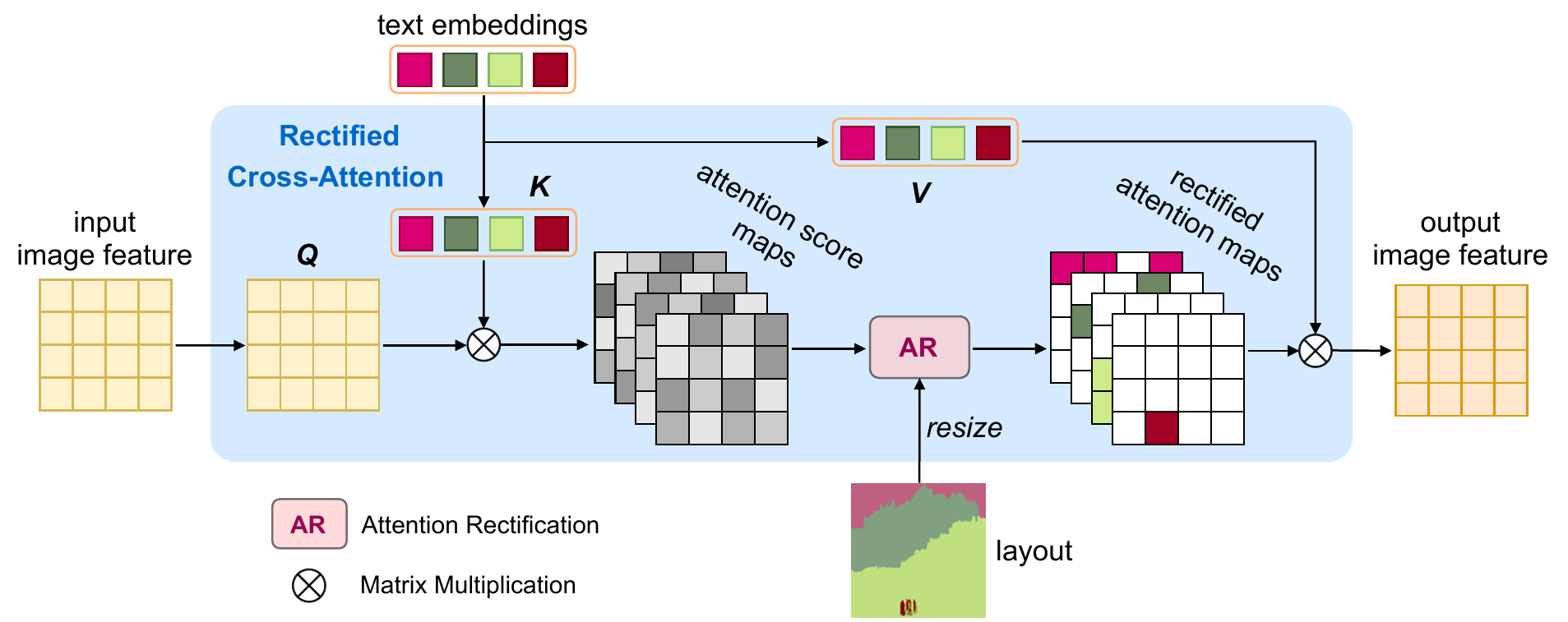}
   %\vspace{-0.1cm}
   \caption{An illustration of Rectified Cross-Attention (RCA). It utilizes the layout to rectify the attention score maps calculated between the text and image tokens. By forcing each text token to affect only pixels in the region specified by the layout, the spatial alignment between the generated image and the given layout is guaranteed.
   $Q$, $K$, $V$ are image queries, text keys, and text values, which are computed through projection layers respectively.
   }
   \label{fig:RCA}
%   \vspace{-0.4cm}
\end{figure*} 
We propose a framework of FLIS (FreestyleNet) by leveraging the pre-trained text-to-image diffusion network: Stable Diffusion~\cite{rombach2022high}. An overview of our method is presented in Figure~\ref{fig:framework}. The original Stable Diffusion adopts a text encoder of CLIP~\cite{radford2021learning} to extract text embeddings from the input text and then injects them into a diffusion model through cross-attention to perform generation. Starting from a noise $x_{T}$, the reverse diffusion is performed using a denoising U-Net inside the model for $T$ time steps to generate the final result $x_{0}$. To empower Stable Diffusion with the ability of synthesizing images on a specified layout, we propose a Rectified Cross-Attention (RCA) to integrate the layout in each cross-attention layer of the U-Net. 
In the following, we will elaborate on 1) how to rectify the pre-trained text-to-image diffusion model in the context of FLIS, and 2) how to further exploit the freestyle capability of the rectified diffusion model for FLIS.

\subsection{Rectifying Diffusion Model}\label{sec:method_sub1}

We propose to rectify the pre-trained diffusion model to facilitate FLIS. The rectification is performed by: 1)~representing semantics by using text embeddings, 2)~injecting semantics to layout through the proposed RCA, and 3)~fine-tuning the diffusion model on layout-based training data.

\noindent\textbf{Representing Semantics.} The rich semantics learned in the pre-trained text-to-image model lie in the textual space. To deploy them for FLIS, we need to represent the desired semantics by texts.
Here we provide a simple yet effective solution. We first describe each semantic (\eg, an object class from COCO-Stuff~\cite{caesar2018coco}) with a word (\eg, ``cat") or several words (\eg, ``teddy bear"), which we call a \textit{concept}.
Then, we stack all \textit{concepts} corresponding to the classes appeared in the input semantic layout (``sky mountain snow person" in Figure~\ref{fig:framework}), and input them into the text encoder. The resulting text embeddings serve as the semantics to perform the following generation. 

\noindent\textbf{Injecting Semantics to Layout.}
Given the text embeddings (\ie, semantics), the next step is to inject the semantics into the input layout. 
However, it is not intuitive how to achieve that on a pre-trained text-to-image diffusion model.

Before diving into our solution, we elaborate the cross-attention layer in this pre-trained model.
The cross-attention layer takes two inputs: text embeddings and image features. It calculates an attention map between them, and outputs a weighted sum on the text values using the weights on the attention map.
Each channel of the attention map is related to a text embedding. For instance, we can find the channel corresponding to a text embedding extracted from the word ``cat''. The spatial layout of this cat in the generated image is similar to this channel, as concluded in \cite{hudson2021generative,hertz2022prompt}. Therefore, by modifying this channel, we can change the shape and location of this cat in the generated image. 
Inspired by this, we propose a Rectified Cross-Attention (RCA) to integrate the input layout into the image generation process. 
The idea of RCA is to rectify the attention maps of each text embedding using the corresponding mask. In this way, it determines the spatial layout of the generated image.

As illustrated in Figure~\ref{fig:RCA}, the input to RCA consists of three parts: input image feature, text embeddings, and layout. RCA computes image queries $Q$, text keys $K$, and text values $V$ through three projection layers, respectively. After that, attention score maps $\mathcal{M}$ are calculated as:
\begin{equation}
  \mathcal{M}=\frac{QK^T}{\sqrt{d}}\in\mathbb{R}^{C\times H\times W},
  \label{eq1}
\end{equation}
where $d$ is the scaling factor that is set as the dimension of the queries and keys, and $C$, $H$, $W$ are the channel number, height, and weight of $\mathcal{M}$, respectively. Next, we utilize the layout to constrain the spatial distribution of the attention maps by rectifying the attention.

We translate the input one-channel semantic layout into a $C$-channel layout, where each channel is a binary map for an individual \textit{concept}. 
For the text embedding corresponding to the \textit{concept}, we use its associated semantic mask as the binary map. 
For other embeddings, \eg, paddings, we set the values on the binary map to $1$. 
To match the spatial size of the attention score maps $\mathcal{M}$, we resize the $C$-channel layout to obtain $L\in\mathbb{R}^{C\times H\times W}$. $\mathcal{M}$ is then rectified by:
\begin{equation}
  \widehat{\mathcal{M}}^{k}_{i,j}= \begin{cases}
  \mathcal{M}^{k}_{i,j}, & L^{k}_{i,j} = 1, \\
  -inf, & L^{k}_{i,j} = 0,
  \end{cases}
\label{eq2}
\end{equation}
where $\mathcal{M}^{k}_{i,j}$ and $L^{k}_{i,j}$ denote the values at position $(i,j)$ in the $k$-th channels of $\mathcal{M}$ and $L$, respectively. $\widehat{\mathcal{M}}$ represents the rectified attention score maps. Thereafter, we can calculate the output image feature $\mathcal{O}$ of this RCA layer by:
\begin{equation}
  \mathcal{O}=\text{softmax}(\widehat{\mathcal{M}})V.
  \label{eq3}
\end{equation}

After applying softmax, the rectified attention map becomes a binary map with the spatial distribution similar to the input layout. Using this map, the model forces each text embedding to affect only pixels in the region specified by the input layout. In other words, the semantics from the text are injected to the layout.

\noindent\textbf{Model Fine-tuning.} Equipped with the proposed RCA, the pre-trained diffusion model can be fine-tuned on any sets of layout-based training data.
The objective for fine-tuning is the same as that for pre-training (image denoising)~\cite{rombach2022high}:
\begin{equation}
  \mathcal{L}(\theta)=\mathbb{E}_{z,l,y,\epsilon\sim\mathcal{N}(0,1),t}\Bigl[\lVert\epsilon-\epsilon_{\theta}(z_{t},t,l,c_{\phi}(y))\rVert_{2}^{2}\Bigr],
  \label{eq4}
\end{equation}
where $z$ is the latent code extracted from the input image, $l$ is the input layout, $y$ is the input text, $\epsilon$ is a noise term, $t$ is the time step, $\epsilon_{\theta}$ is the denoising U-Net, $z_{t}$ is the noisy version of $z$ at time $t$, and $c_{\phi}$ is the text encoder. We merely fine-tune the denoising U-Net while keeping the text encoder and the autoencoder of Stable Diffusion frozen. We refer readers to the original paper of Stable Diffusion~\cite{rombach2022high} for the training details.

\subsection{Generating Freestyle Images}\label{sec:method_sub2}
\begin{figure*}[t]
   \centering
   \includegraphics[width=1\linewidth]{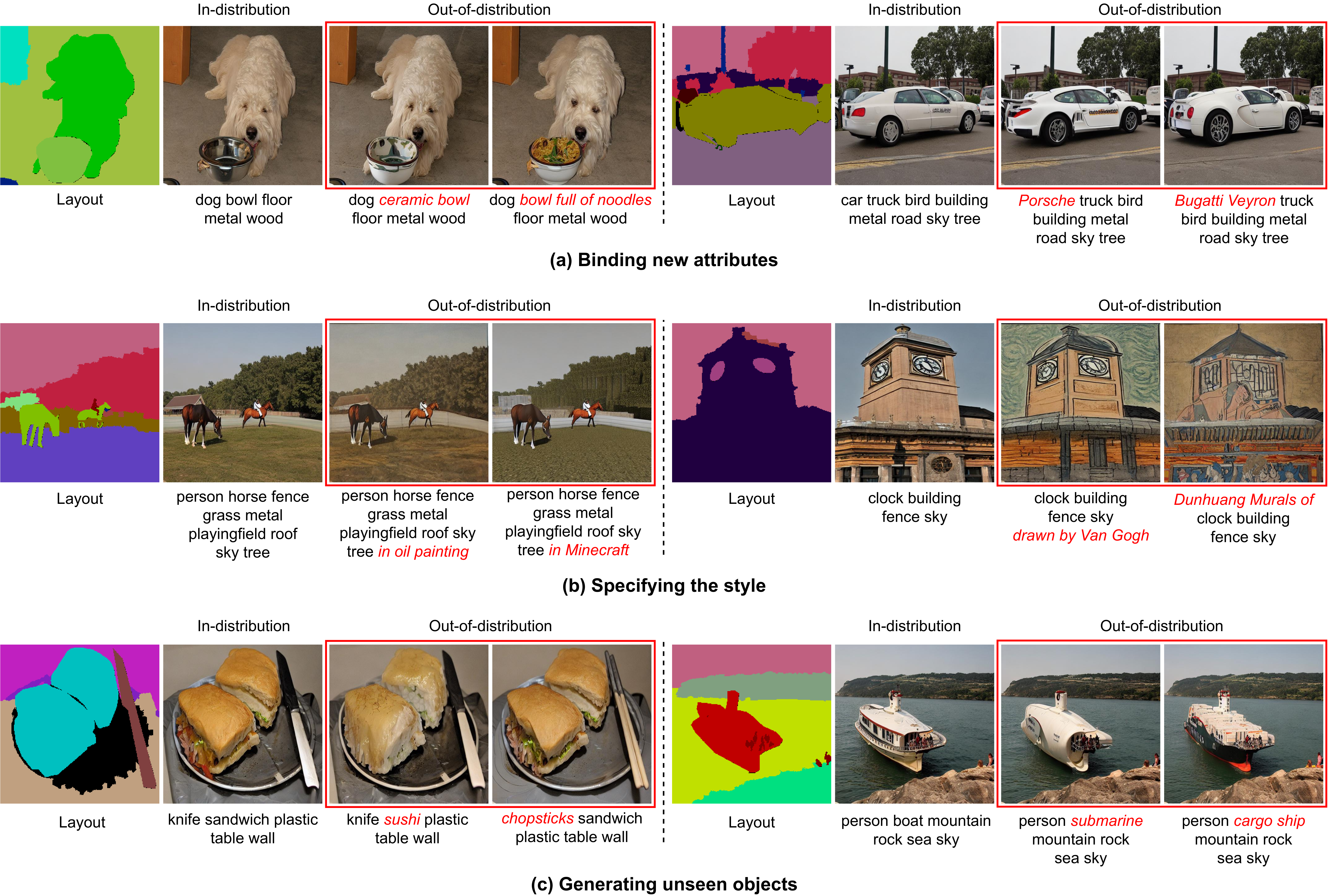}
   %\vspace{-0.1cm}
   \caption{Example results and capabilities from our FreestyleNet in performing freestyle layout-to-image synthesis (FLIS). Zoom in for a better view.
   }
   \label{fig:my_results}
%   \vspace{-0.4cm}
\end{figure*} 
The fine-tuned diffusion model is endowed with the ability to synthesize images given both layouts and texts. By describing the desired semantics in the input text, we generalize the model by excavating its pre-trained knowledge to perform freestyle layout-to-image synthesis. Here we showcase three capabilities of the proposed model for FLIS as well as the mechanisms behind them.

\noindent\textbf{Binding New Attributes.} Having a layout that contains an object of interest (\eg, cat), one may further want to generate a specific species of cat (\eg, tabby cat). In this scenario, we first replace ``cat" with ``tabby cat" in the text input. Then, following \cref{eq2}, we adopt the binary map of the input layout $L$ (corresponding to the region you want to place this tabby cat) to rectify the two attention score maps, which are related to ``tabby" and ``cat" respectively.

\noindent\textbf{Specifying the Style.} We can render a given image in the desired style by adding a description (\eg, ``an ink painting of") to the input text. Since we aim to specify the global style of the generated image, the attention score maps corresponding to these added words are not rectified.

\noindent\textbf{Generating Unseen Objects.} Benefiting from the general knowledge learned from the pre-training phase, our model is able to generate objects from unseen classes. This is achieved by simply describing the object in the text input and rectifying the corresponding attention map through \cref{eq2} with the layout related to this object.
\section{Experiments}\label{sec:exper}

\subsection{Experimental Settings}

\noindent\textbf{Datasets.} We conduct experiments on two challenging datasets: COCO-Stuff~\cite{caesar2018coco} and ADE20K~\cite{zhou2017scene}. COCO-Stuff consists of 118,287 training and 5,000 validation images, which are annotated with 182 semantic classes. ADE20K has 20,210 images for training and 2,000 for validation, including 150 semantic classes.

\noindent\textbf{Implementation Details.} We adopt the original training settings of Stable Diffusion~\cite{rombach2022high} to optimize our model, unless otherwise noted. The base learning rate is set to $10^{-6}$. We use 50 PLMS~\cite{liu2022pseudo} sampling steps with a classifier-free guidance~\cite{ho2022classifier} scale of 2. All the experiments are conducted on a single NVIDIA A100 GPU. 

\subsection{Qualitative Evaluation on FLIS}

\begin{figure*}[t]
   \centering
   \includegraphics[width=0.98\linewidth]{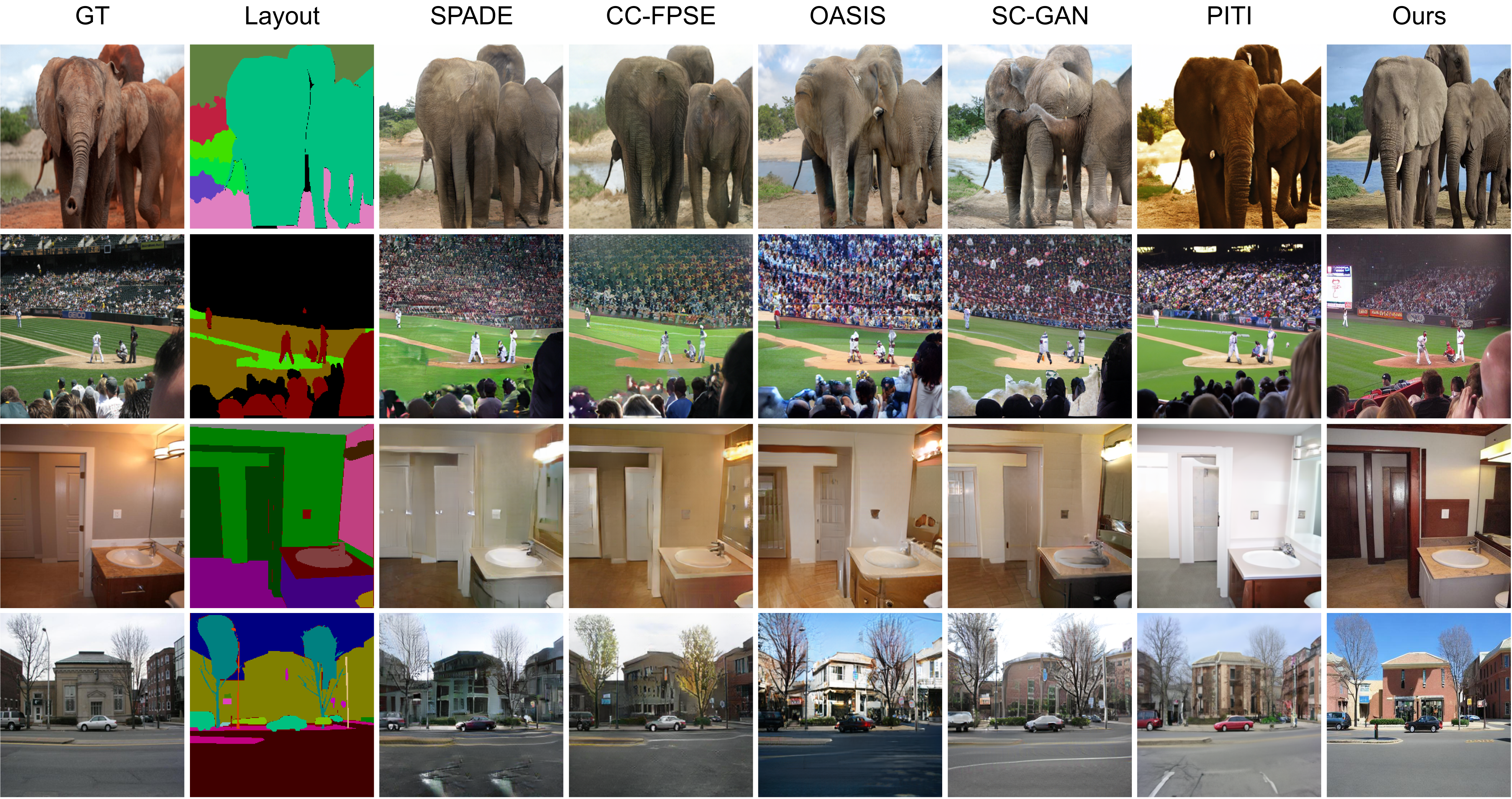}
   %\vspace{-0.1cm}
   \caption{Qualitative comparison results with the state-of-the-art layout-to-image synthesis methods on COCO-Stuff (top 2 rows) and ADE20K (bottom 2 rows). Note that our FreestyleNet takes text and layout as input while other methods adopt only the layout.
   }
   \label{fig:comparison}
%   \vspace{-0.4cm}
\end{figure*} 
In order to demonstrate the viability of our FreestyleNet for freestyle layout-to-image synthesis (FLIS), we showcase some synthesis results covering a variety of scenarios. As presented in Figures~\ref{fig:teaser} and \ref{fig:my_results}, the proposed model is capable of but not limited to 1) binding new attributes to the objects, 2) specifying the styles for the synthesized images, and 3) generating unseen objects. The model we use to generate these images is fine-tuned on COCO-Stuff. For each sample here, we conduct two kinds of layout-to-image synthesis, \ie, in-distribution and out-of-distribution. For the former, the text input only contains the semantics that are defined in the 182 classes of COCO-Stuff. While for the out-of-distribution synthesis (depicted with a red box), we apply some words with a wide range of new semantics in the text input. Our method consistently produces high-fidelity images that faithfully reveal the novel semantics described in the text while conforming to the given layouts. Some impressive features enabled by our approach such as generating a ``Bugatti Veyron" or hornless ``unicorn" and putting the scene ``in Minecraft" break the in-distribution limit, allowing the users to create the images in a \textit{freestyle} way\footnote{A comparison to a concurrent work ControlNet~\cite{zhang2023adding} is provided on our GitHub page.}.

\subsection{Comparison with LIS Baselines}
We compare our method against the state-of-the-art LIS baselines including Pix2PixHD~\cite{wang2018high}, SPADE~\cite{park2019semantic}, CC-FPSE~\cite{liu2019learning}, OASIS~\cite{sushko2020you}, SC-GAN~\cite{wang2021image}, and PITI~\cite{wang2022pretraining}. \emph{Notably, since generating unseen semantics is beyond the reach of these methods, we conduct the comparison under the in-distribution setting.} While other methods use only one input modality (layout), the proposed model adopts two (text and layout). We follow the standard test settings to evaluate all methods on COCO-Stuff and ADE20K. Fr\'echet Inception Distance (FID)~\cite{heusel2017gans} and mean Intersection-over-Union (mIoU) are adopted to assess the realism of the generated images and the semantic alignment between the generated images and the input layouts, respectively.

\begin{table}
\setlength{\tabcolsep}{4pt}
  \centering
  \begin{tabular}{ccccccc}
    \toprule[0.8pt]
    \multirow{2.5}*{Method} & \textbf{} & \multicolumn{2}{c}{COCO-Stuff} & \textbf{} & \multicolumn{2}{c}{ADE20K} \\
    \cmidrule[0.5pt]{3-4}
    \cmidrule[0.5pt]{6-7}
    ~ & \textbf{} & FID$\downarrow$ & mIoU$\uparrow$ & \textbf{} & FID$\downarrow$ & mIoU$\uparrow$ \\
    \midrule[0.5pt]
    Pix2PixHD~\cite{wang2018high} & \textbf{} & 111.5 & 14.6 & \textbf{} & 81.8 & 20.3 \\
    SPADE~\cite{park2019semantic} & \textbf{} & 22.6 & 37.4 & \textbf{} & 33.9 & 38.5 \\
    CC-FPSE~\cite{liu2019learning} & \textbf{} & 19.2 & 41.6 & \textbf{} & 31.7 & 43.7 \\
    OASIS~\cite{sushko2020you} & \textbf{} & 17.0 & \textbf{44.1} & \textbf{} & 28.3 & \textbf{48.8} \\
    SC-GAN~\cite{wang2021image} & \textbf{} & 18.1 & 42.0 & \textbf{} & 29.3 & 45.2 \\
    PITI~\cite{wang2022pretraining} & \textbf{} & 16.1 & 34.1 & \textbf{} & 27.9 & 29.4 \\
    FreestyleNet (ours) & \textbf{} & \textbf{14.4} & 40.7 & \textbf{} & \textbf{25.0} & 41.9 \\
    \bottomrule[0.8pt]
  \end{tabular}
  %\vspace{2mm}
  \caption{Quantitative comparison results with the state-of-the-art layout-to-image synthesis methods. Note that our model takes text and layout as input while other methods adopt only the layout.}
  \label{tab:comparison}
  \vspace{-0.35cm}
\end{table}

Quantitative comparison results are reported in Table~\ref{tab:comparison}. The proposed model outperforms all the competing methods in terms of the FID metric, indicating the high visual quality of the images synthesized by our method. The most recent method, PITI, also exhibits superior generation quality by leveraging a pre-trained diffusion model. However, it implicitly learns to map the input layouts into the space of the pre-trained text encoder, resulting in clearly worse spatial alignment (reflected in low mIoU values) than ours. Although our method does not suffer from this problem by explicitly constraining the spatial layout of image features through RCA, it obtains less preferred results in mIoU compared to some of the other LIS methods. We argue that mIoU is not fully suitable to evaluate the proposed FLIS method. While the text input does not contain unseen semantics under the in-distribution setting, our method may sometimes generate more general semantics that stem from the pre-trained knowledge (\eg., the wall behind the switch in the 3rd row of Figure~\ref{fig:comparison}), leading to mispredictions of class labels by the segmentation model used to calculate mIoU. In addition, our goal is to synthesize images that are \emph{spatially}, not semantically, aligned with the input layout. As later shown in the visual comparison, our generated images have a strong spatial alignment with the input layouts.

Qualitative comparison results are shown in Figure~\ref{fig:comparison}. Compared to the other baselines, our method consistently produces sharper images with fine details, which is consistent with the quantitative evaluation results.

For a fair comparison, we replace PITI's backbone GLIDE~\cite{nichol2021glide} with Stable Diffusion. The comparison results (see Figure~\ref{fig:piti-wsd} and Table~\ref{tab:fid_comp} in the supplementary) indicate that the clearly improved spatial alignment are due to RCA, and the image quality is also enhanced by our framework.

\subsection{Effectiveness of RCA}
\begin{figure}[t]
   \centering
   \includegraphics[width=0.95\linewidth]{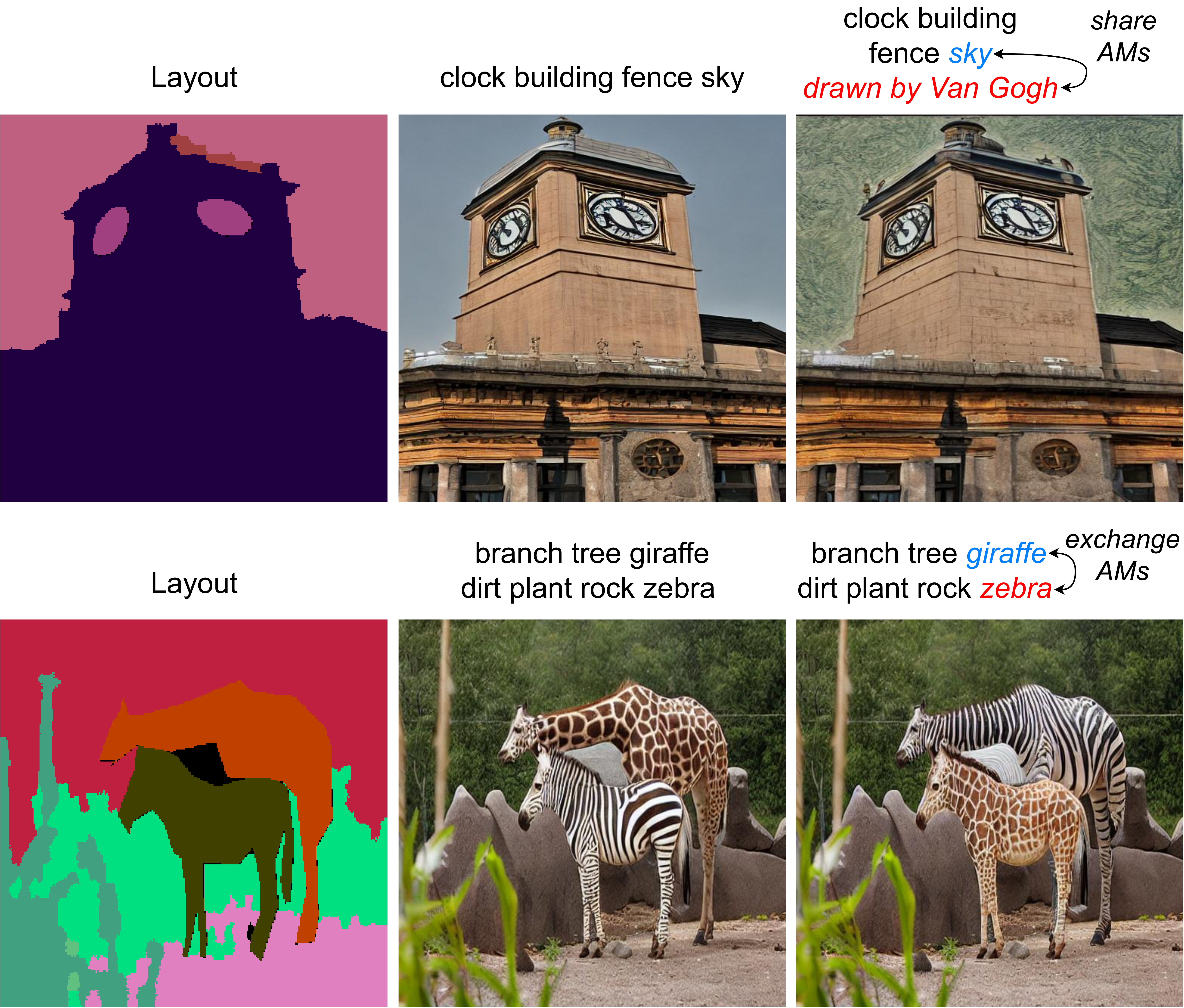}
   %\vspace{-0.1cm}
  \caption{Effectiveness of the proposed RCA. AM represents attention maps in each RCA. By sharing/exchanging the attention maps rectified by RCA between different semantics, our method succeeds in local style assignment/appearance swapping.}
   \label{fig:effec_RCA}
  \vspace{-0.2cm}
\end{figure} 

To further validate the effectiveness of the proposed RCA, we show two representative examples in Figure~\ref{fig:effec_RCA}. In the first row, if we share the rectified attention maps corresponding to ``sky" with those associated with ``drawn by Van Gogh" when specifying the image style, only the sky will be rendered in the style of Van Gogh. The second row showcases that our model is also capable of swapping the appearance of two objects by exchanging their corresponding rectified attention maps. Both of these two cases demonstrate that RCA can force the semantics described by the text to appear in the region specified by the layout.

\subsection{Limitations}
\begin{figure}[t]
   \centering
   \includegraphics[width=0.95\linewidth]{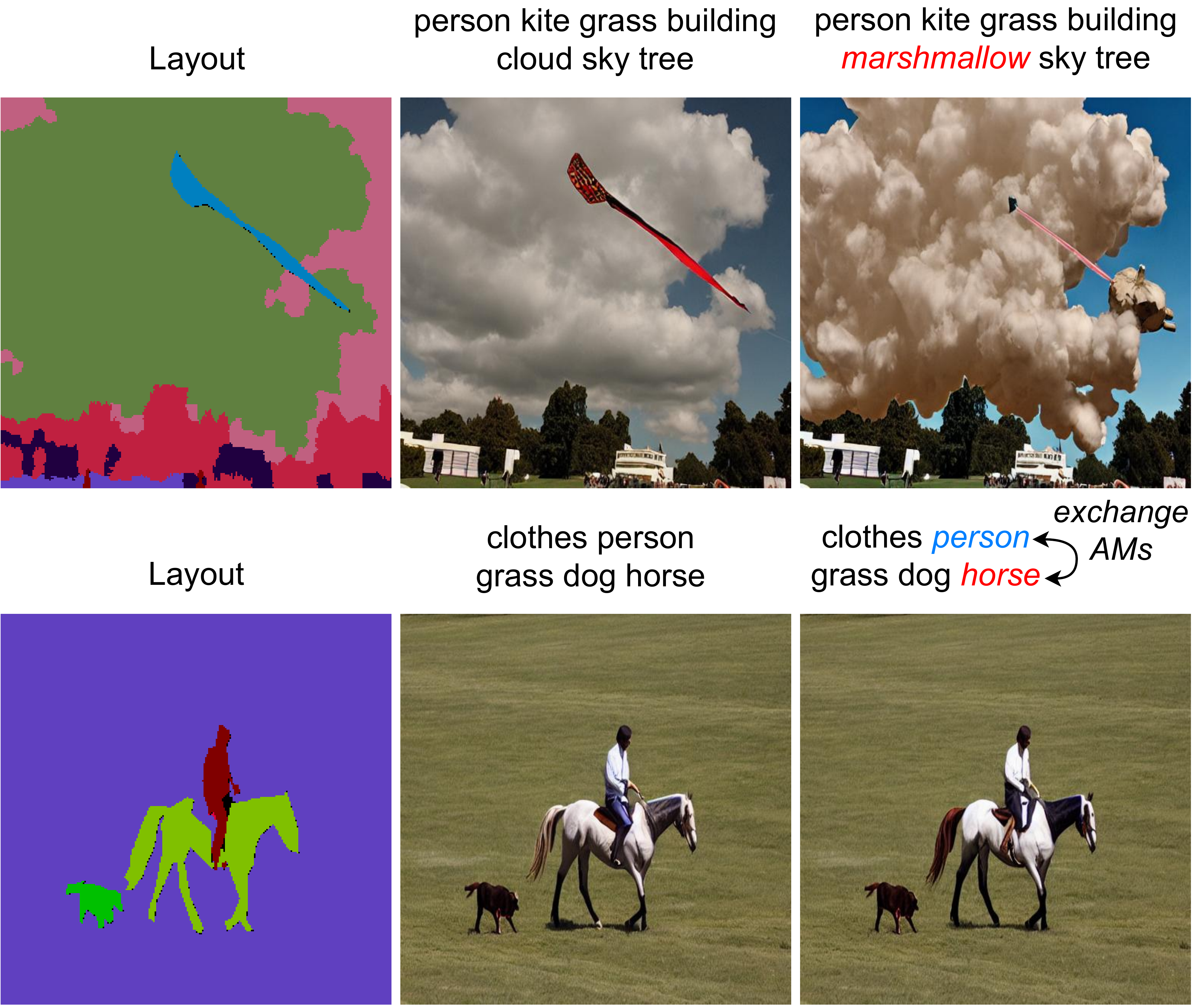}
   %\vspace{-0.1cm}
  \caption{Failure cases. AM represents attention maps in each RCA. Our method may struggle with generating some rare semantics. More results are provided in the supplementary material.}
   \label{fig:failure_case}
  \vspace{-0.2cm}
\end{figure} 

We find several failure cases of our method. As presented in Figure~\ref{fig:failure_case}, our model cannot produce satisfactory results (distortion in kite and failed appearance swap) when the specified semantics or scenes are counterfactual (\eg, marshmallow in the sky and horse riding a person).
One possible reason is that the diffusion model itself does not learn sufficient knowledge of these rarely appeared or unreasonable semantics. Another reason could be the fine-tuning on downstream datasets, likely resulting in some damage to pre-trained knowledge.
Our model cannot work as intended if given contradictory layout and prompt (\eg, the unicorn in Figure~\ref{fig:teaser} is hornless). A feasible solution is to relax the layout constraints, \eg, by increasing the rectified attention score when $L_{i,j}^{k}=0$ (indicating regions outside the $k$-th semantic) in \cref{eq2}.
Besides, our method needs to have a user-defined set of category names, which could be expensive to collect in case of long-tailed datasets.

\section{Conclusions}\label{sec:conclus}

In this paper, we introduce a new task called freestyle layout-to-image synthesis (FLIS), which aims to generate unseen semantics described by text onto a given layout. To that end, we propose to leverage the generative prior from pre-trained text-to-image diffusion models. After inserting the proposed Rectified Cross-Attention (RCA) module, the pre-trained model is armed with the FLIS capability, breaking the in-distribution limit that hinders the applicability of previous LIS models. We demonstrate a plethora of applications by translating a single layout into high-fidelity images with a variety of novel semantics using our FreestyleNet.

\noindent\textbf{Acknowledgments.} \small{This work was funded by the Singapore MOE Academic Research Fund Tier 1 grant (MSS21C002). It was also supported by ``the Fundamental Research Funds for the Central Universities'', 111 Project, China under Grants B07022 and (Sheitc) 150633, and Shanghai Key Laboratory of Digital Media Processing and Transmissions, China. It was also funded by the A*STAR under its AME YIRG Grant (Project No.A20E6c0101).}

%%%%%%%%% REFERENCES
{\small
\bibliographystyle{ieee_fullname}
\bibliography{main}

\begin{thebibliography}{10}\itemsep=-1pt

\bibitem{alaniz2022semantic}
Stephan Alaniz, Thomas Hummel, and Zeynep Akata.
\newblock Semantic image synthesis with semantically coupled vq-model.
\newblock {\em arXiv preprint arXiv:2209.02536}, 2022.

\bibitem{brock2018large}
Andrew Brock, Jeff Donahue, and Karen Simonyan.
\newblock Large scale gan training for high fidelity natural image synthesis.
\newblock {\em arXiv preprint arXiv:1809.11096}, 2018.

\bibitem{caesar2018coco}
Holger Caesar, Jasper Uijlings, and Vittorio Ferrari.
\newblock Coco-stuff: Thing and stuff classes in context.
\newblock In {\em CVPR}, pages 1209--1218, 2018.

\bibitem{chen2017photographic}
Qifeng Chen and Vladlen Koltun.
\newblock Photographic image synthesis with cascaded refinement networks.
\newblock In {\em ICCV}, pages 1511--1520, 2017.

\bibitem{cordts2016cityscapes}
Marius Cordts, Mohamed Omran, Sebastian Ramos, Timo Rehfeld, Markus Enzweiler,
  Rodrigo Benenson, Uwe Franke, Stefan Roth, and Bernt Schiele.
\newblock The cityscapes dataset for semantic urban scene understanding.
\newblock In {\em CVPR}, pages 3213--3223, 2016.

\bibitem{dhariwal2021diffusion}
Prafulla Dhariwal and Alexander Nichol.
\newblock Diffusion models beat gans on image synthesis.
\newblock In {\em NeurIPS}, pages 8780--8794, 2021.

\bibitem{ding2022cogview2}
Ming Ding, Wendi Zheng, Wenyi Hong, and Jie Tang.
\newblock Cogview2: Faster and better text-to-image generation via hierarchical
  transformers.
\newblock {\em arXiv preprint arXiv:2204.14217}, 2022.

\bibitem{fan2022frido}
Wan-Cyuan Fan, Yen-Chun Chen, Dongdong Chen, Yu Cheng, Lu Yuan, and
  Yu-Chiang~Frank Wang.
\newblock Frido: Feature pyramid diffusion for complex scene image synthesis.
\newblock In {\em AAAI}, 2023.

\bibitem{gafni2022make}
Oran Gafni, Adam Polyak, Oron Ashual, Shelly Sheynin, Devi Parikh, and Yaniv
  Taigman.
\newblock Make-a-scene: Scene-based text-to-image generation with human priors.
\newblock In {\em ECCV}, pages 89--106, 2022.

\bibitem{gal2022image}
Rinon Gal, Yuval Alaluf, Yuval Atzmon, Or Patashnik, Amit~H Bermano, Gal
  Chechik, and Daniel Cohen-Or.
\newblock An image is worth one word: Personalizing text-to-image generation
  using textual inversion.
\newblock {\em arXiv preprint arXiv:2208.01618}, 2022.

\bibitem{DBLP:conf/nips/GoodfellowPMXWOCB14}
Ian~J. Goodfellow, Jean Pouget{-}Abadie, Mehdi Mirza, Bing Xu, David
  Warde{-}Farley, Sherjil Ozair, Aaron~C. Courville, and Yoshua Bengio.
\newblock Generative adversarial nets.
\newblock In {\em NeurIPS}, pages 2672--2680, 2014.

\bibitem{hertz2022prompt}
Amir Hertz, Ron Mokady, Jay Tenenbaum, Kfir Aberman, Yael Pritch, and Daniel
  Cohen-Or.
\newblock Prompt-to-prompt image editing with cross attention control.
\newblock {\em arXiv preprint arXiv:2208.01626}, 2022.

\bibitem{heusel2017gans}
Martin Heusel, Hubert Ramsauer, Thomas Unterthiner, Bernhard Nessler, and Sepp
  Hochreiter.
\newblock Gans trained by a two time-scale update rule converge to a local nash
  equilibrium.
\newblock In {\em NeurIPS}, 2017.

\bibitem{hinz2019generating}
Tobias Hinz, Stefan Heinrich, and Stefan Wermter.
\newblock Generating multiple objects at spatially distinct locations.
\newblock {\em arXiv preprint arXiv:1901.00686}, 2019.

\bibitem{ho2020denoising}
Jonathan Ho, Ajay Jain, and Pieter Abbeel.
\newblock Denoising diffusion probabilistic models.
\newblock In {\em NeurIPS}, pages 6840--6851, 2020.

\bibitem{ho2022classifier}
Jonathan Ho and Tim Salimans.
\newblock Classifier-free diffusion guidance.
\newblock {\em arXiv preprint arXiv:2207.12598}, 2022.

\bibitem{hong2018inferring}
Seunghoon Hong, Dingdong Yang, Jongwook Choi, and Honglak Lee.
\newblock Inferring semantic layout for hierarchical text-to-image synthesis.
\newblock In {\em CVPR}, pages 7986--7994, 2018.

\bibitem{huang2022multimodal}
Xun Huang, Arun Mallya, Ting-Chun Wang, and Ming-Yu Liu.
\newblock Multimodal conditional image synthesis with product-of-experts gans.
\newblock In {\em ECCV}, pages 91--109, 2022.

\bibitem{hudson2021generative}
Drew~A Hudson and Larry Zitnick.
\newblock Generative adversarial transformers.
\newblock In {\em ICML}, pages 4487--4499. PMLR, 2021.

\bibitem{isola2017image}
Phillip Isola, Jun-Yan Zhu, Tinghui Zhou, and Alexei~A Efros.
\newblock Image-to-image translation with conditional adversarial networks.
\newblock In {\em CVPR}, pages 1125--1134, 2017.

\bibitem{karras2019style}
Tero Karras, Samuli Laine, and Timo Aila.
\newblock A style-based generator architecture for generative adversarial
  networks.
\newblock In {\em CVPR}, pages 4401--4410, 2019.

\bibitem{kawar2022imagic}
Bahjat Kawar, Shiran Zada, Oran Lang, Omer Tov, Huiwen Chang, Tali Dekel, Inbar
  Mosseri, and Michal Irani.
\newblock Imagic: Text-based real image editing with diffusion models.
\newblock {\em arXiv preprint arXiv:2210.09276}, 2022.

\bibitem{liang2022layout}
Jiadong Liang, Wenjie Pei, and Feng Lu.
\newblock Layout-bridging text-to-image synthesis.
\newblock {\em arXiv preprint arXiv:2208.06162}, 2022.

\bibitem{liu2022pseudo}
Luping Liu, Yi Ren, Zhijie Lin, and Zhou Zhao.
\newblock Pseudo numerical methods for diffusion models on manifolds.
\newblock {\em arXiv preprint arXiv:2202.09778}, 2022.

\bibitem{liu2019learning}
Xihui Liu, Guojun Yin, Jing Shao, Xiaogang Wang, et~al.
\newblock Learning to predict layout-to-image conditional convolutions for
  semantic image synthesis.
\newblock In {\em NeurIPS}, 2019.

\bibitem{lugmayr2022repaint}
Andreas Lugmayr, Martin Danelljan, Andres Romero, Fisher Yu, Radu Timofte, and
  Luc Van~Gool.
\newblock Repaint: Inpainting using denoising diffusion probabilistic models.
\newblock In {\em CVPR}, pages 11461--11471, 2022.

\bibitem{lv2021learning}
Zhengyao Lv, Xiaoming Li, Xin Li, Fu Li, Tianwei Lin, Dongliang He, and
  Wangmeng Zuo.
\newblock Learning semantic person image generation by region-adaptive
  normalization.
\newblock In {\em CVPR}, pages 10806--10815, 2021.

\bibitem{lv2022semantic}
Zhengyao Lv, Xiaoming Li, Zhenxing Niu, Bing Cao, and Wangmeng Zuo.
\newblock Semantic-shape adaptive feature modulation for semantic image
  synthesis.
\newblock In {\em CVPR}, pages 11214--11223, 2022.

\bibitem{nichol2021glide}
Alex Nichol, Prafulla Dhariwal, Aditya Ramesh, Pranav Shyam, Pamela Mishkin,
  Bob McGrew, Ilya Sutskever, and Mark Chen.
\newblock Glide: Towards photorealistic image generation and editing with
  text-guided diffusion models.
\newblock {\em arXiv preprint arXiv:2112.10741}, 2021.

\bibitem{park2019semantic}
Taesung Park, Ming-Yu Liu, Ting-Chun Wang, and Jun-Yan Zhu.
\newblock Semantic image synthesis with spatially-adaptive normalization.
\newblock In {\em CVPR}, pages 2337--2346, 2019.

\bibitem{qi2018semi}
Xiaojuan Qi, Qifeng Chen, Jiaya Jia, and Vladlen Koltun.
\newblock Semi-parametric image synthesis.
\newblock In {\em CVPR}, pages 8808--8816, 2018.

\bibitem{radford2021learning}
Alec Radford, Jong~Wook Kim, Chris Hallacy, Aditya Ramesh, Gabriel Goh,
  Sandhini Agarwal, Girish Sastry, Amanda Askell, Pamela Mishkin, Jack Clark,
  et~al.
\newblock Learning transferable visual models from natural language
  supervision.
\newblock In {\em ICML}, pages 8748--8763. PMLR, 2021.

\bibitem{ramesh2022hierarchical}
Aditya Ramesh, Prafulla Dhariwal, Alex Nichol, Casey Chu, and Mark Chen.
\newblock Hierarchical text-conditional image generation with clip latents.
\newblock {\em arXiv preprint arXiv:2204.06125}, 2022.

\bibitem{reed2016learning}
Scott~E Reed, Zeynep Akata, Santosh Mohan, Samuel Tenka, Bernt Schiele, and
  Honglak Lee.
\newblock Learning what and where to draw.
\newblock In {\em NeurIPS}, 2016.

\bibitem{rombach2022high}
Robin Rombach, Andreas Blattmann, Dominik Lorenz, Patrick Esser, and Bj{\"o}rn
  Ommer.
\newblock High-resolution image synthesis with latent diffusion models.
\newblock In {\em CVPR}, pages 10684--10695, 2022.

\bibitem{ruiz2022dreambooth}
Nataniel Ruiz, Yuanzhen Li, Varun Jampani, Yael Pritch, Michael Rubinstein, and
  Kfir Aberman.
\newblock Dreambooth: Fine tuning text-to-image diffusion models for
  subject-driven generation.
\newblock {\em arXiv preprint arXiv:2208.12242}, 2022.

\bibitem{saharia2022photorealistic}
Chitwan Saharia, William Chan, Saurabh Saxena, Lala Li, Jay Whang, Emily
  Denton, Seyed Kamyar~Seyed Ghasemipour, Burcu~Karagol Ayan, S~Sara Mahdavi,
  Rapha~Gontijo Lopes, et~al.
\newblock Photorealistic text-to-image diffusion models with deep language
  understanding.
\newblock {\em arXiv preprint arXiv:2205.11487}, 2022.

\bibitem{oasis}
Edgar Sch{\"o}nfeld, Vadim Sushko, Dan Zhang, Juergen Gall, Bernt Schiele, and
  Anna Khoreva.
\newblock You only need adversarial supervision for semantic image synthesis.
\newblock In {\em ICLR}, 2021.

\bibitem{schuhmann2022laion}
Christoph Schuhmann, Romain Beaumont, Richard Vencu, Cade Gordon, Ross
  Wightman, Mehdi Cherti, Theo Coombes, Aarush Katta, Clayton Mullis, Mitchell
  Wortsman, et~al.
\newblock Laion-5b: An open large-scale dataset for training next generation
  image-text models.
\newblock {\em arXiv preprint arXiv:2210.08402}, 2022.

\bibitem{shi2022retrieval}
Yupeng Shi, Xiao Liu, Yuxiang Wei, Zhongqin Wu, and Wangmeng Zuo.
\newblock Retrieval-based spatially adaptive normalization for semantic image
  synthesis.
\newblock In {\em CVPR}, pages 11224--11233, 2022.

\bibitem{sohl2015deep}
Jascha Sohl-Dickstein, Eric Weiss, Niru Maheswaranathan, and Surya Ganguli.
\newblock Deep unsupervised learning using nonequilibrium thermodynamics.
\newblock In {\em ICML}, pages 2256--2265. PMLR, 2015.

\bibitem{song2020denoising}
Jiaming Song, Chenlin Meng, and Stefano Ermon.
\newblock Denoising diffusion implicit models.
\newblock {\em arXiv preprint arXiv:2010.02502}, 2020.

\bibitem{song2020improved}
Yang Song and Stefano Ermon.
\newblock Improved techniques for training score-based generative models.
\newblock In {\em NeurIPS}, pages 12438--12448, 2020.

\bibitem{song2020score}
Yang Song, Jascha Sohl-Dickstein, Diederik~P Kingma, Abhishek Kumar, Stefano
  Ermon, and Ben Poole.
\newblock Score-based generative modeling through stochastic differential
  equations.
\newblock {\em arXiv preprint arXiv:2011.13456}, 2020.

\bibitem{sushko2020you}
Vadim Sushko, Edgar Sch{\"o}nfeld, Dan Zhang, Juergen Gall, Bernt Schiele, and
  Anna Khoreva.
\newblock You only need adversarial supervision for semantic image synthesis.
\newblock {\em arXiv preprint arXiv:2012.04781}, 2020.

\bibitem{tan2021efficient}
Zhentao Tan, Dongdong Chen, Qi Chu, Menglei Chai, Jing Liao, Mingming He, Lu
  Yuan, Gang Hua, and Nenghai Yu.
\newblock Efficient semantic image synthesis via class-adaptive normalization.
\newblock {\em TPAMI}, 2021.

\bibitem{tan2022semantic}
Zhentao Tan, Qi Chu, Menglei Chai, Dongdong Chen, Jing Liao, Qiankun Liu, Bin
  Liu, Gang Hua, and Nenghai Yu.
\newblock Semantic probability distribution modeling for diverse semantic image
  synthesis.
\newblock {\em TPAMI}, 2022.

\bibitem{tang2020dual}
Hao Tang, Song Bai, and Nicu Sebe.
\newblock Dual attention gans for semantic image synthesis.
\newblock In {\em ACMMM}, pages 1994--2002, 2020.

\bibitem{wang2022pretraining}
Tengfei Wang, Ting Zhang, Bo Zhang, Hao Ouyang, Dong Chen, Qifeng Chen, and
  Fang Wen.
\newblock Pretraining is all you need for image-to-image translation.
\newblock {\em arXiv preprint arXiv:2205.12952}, 2022.

\bibitem{wang2018high}
Ting-Chun Wang, Ming-Yu Liu, Jun-Yan Zhu, Andrew Tao, Jan Kautz, and Bryan
  Catanzaro.
\newblock High-resolution image synthesis and semantic manipulation with
  conditional gans.
\newblock In {\em CVPR}, pages 8798--8807, 2018.

\bibitem{wang2022semantic}
Weilun Wang, Jianmin Bao, Wengang Zhou, Dongdong Chen, Dong Chen, Lu Yuan, and
  Houqiang Li.
\newblock Semantic image synthesis via diffusion models.
\newblock {\em arXiv preprint arXiv:2207.00050}, 2022.

\bibitem{wang2022sprompt}
Yabin Wang, Zhiwu Huang, and Xiaopeng Hong.
\newblock S-prompts learning with pre-trained transformers: An occam's razor
  for domain incremental learning.
\newblock In {\em NeurIPS}, 2022.

\bibitem{wang2021image}
Yi Wang, Lu Qi, Ying-Cong Chen, Xiangyu Zhang, and Jiaya Jia.
\newblock Image synthesis via semantic composition.
\newblock In {\em ICCV}, pages 13749--13758, 2021.

\bibitem{xu2018attngan}
Tao Xu, Pengchuan Zhang, Qiuyuan Huang, Han Zhang, Zhe Gan, Xiaolei Huang, and
  Xiaodong He.
\newblock Attngan: Fine-grained text to image generation with attentional
  generative adversarial networks.
\newblock In {\em CVPR}, pages 1316--1324, 2018.

\bibitem{yu2022scaling}
Jiahui Yu, Yuanzhong Xu, Jing~Yu Koh, Thang Luong, Gunjan Baid, Zirui Wang,
  Vijay Vasudevan, Alexander Ku, Yinfei Yang, Burcu~Karagol Ayan, et~al.
\newblock Scaling autoregressive models for content-rich text-to-image
  generation.
\newblock {\em arXiv preprint arXiv:2206.10789}, 2022.

\bibitem{zhang2021cross}
Han Zhang, Jing~Yu Koh, Jason Baldridge, Honglak Lee, and Yinfei Yang.
\newblock Cross-modal contrastive learning for text-to-image generation.
\newblock In {\em CVPR}, pages 833--842, 2021.

\bibitem{zhang2017stackgan}
Han Zhang, Tao Xu, Hongsheng Li, Shaoting Zhang, Xiaogang Wang, Xiaolei Huang,
  and Dimitris~N Metaxas.
\newblock Stackgan: Text to photo-realistic image synthesis with stacked
  generative adversarial networks.
\newblock In {\em ICCV}, pages 5907--5915, 2017.

\bibitem{zhang2023adding}
Lvmin Zhang and Maneesh Agrawala.
\newblock Adding conditional control to text-to-image diffusion models.
\newblock {\em arXiv preprint arXiv:2302.05543}, 2023.

\bibitem{zhang2018unreasonable}
Richard Zhang, Phillip Isola, Alexei~A Efros, Eli Shechtman, and Oliver Wang.
\newblock The unreasonable effectiveness of deep features as a perceptual
  metric.
\newblock In {\em CVPR}, pages 586--595, 2018.

\bibitem{zhang2021ufc}
Zhu Zhang, Jianxin Ma, Chang Zhou, Rui Men, Zhikang Li, Ming Ding, Jie Tang,
  Jingren Zhou, and Hongxia Yang.
\newblock Ufc-bert: Unifying multi-modal controls for conditional image
  synthesis.
\newblock In {\em NeurIPS}, pages 27196--27208, 2021.

\bibitem{zhou2017scene}
Bolei Zhou, Hang Zhao, Xavier Puig, Sanja Fidler, Adela Barriuso, and Antonio
  Torralba.
\newblock Scene parsing through ade20k dataset.
\newblock In {\em CVPR}, pages 633--641, 2017.

\bibitem{zhou2022learning}
Kaiyang Zhou, Jingkang Yang, Chen~Change Loy, and Ziwei Liu.
\newblock Learning to prompt for vision-language models.
\newblock {\em IJCV}, 130(9):2337--2348, 2022.

\bibitem{zhu2022label}
Junchen Zhu, Lianli Gao, Jingkuan Song, Yuan-Fang Li, Feng Zheng, Xuelong Li,
  and Heng~Tao Shen.
\newblock Label-guided generative adversarial network for realistic image
  synthesis.
\newblock {\em TPAMI}, 2022.

\bibitem{zhu2020sean}
Peihao Zhu, Rameen Abdal, Yipeng Qin, and Peter Wonka.
\newblock Sean: Image synthesis with semantic region-adaptive normalization.
\newblock In {\em CVPR}, pages 5104--5113, 2020.

\bibitem{zhu2020semantically}
Zhen Zhu, Zhiliang Xu, Ansheng You, and Xiang Bai.
\newblock Semantically multi-modal image synthesis.
\newblock In {\em CVPR}, pages 5467--5476, 2020.

\end{thebibliography}
}

\clearpage
\beginsupp

\noindent
{\Large {\textbf{Supplementary materials}}}
\\

This supplementary material includes an extensive description of Cross-Attention (CA) (\S\ref{sec_CA}), the algorithm of Rectified Cross-Attention (RCA) (\S\ref{sec_algorithm}), additional implementation details (\S\ref{sec_additional_implementation_details}), more qualitative results on freestyle layout-to-image synthesis (FLIS) (\S\ref{sec_more_qualitative_results_on_FLIS}), more comparisons with layout-to-image synthesis (LIS) baselines (\S\ref{sec_more_comparisons_with_LIS_baselines}), the diversity evaluation (\S\ref{sec_diversity_evaluation}), discussions about the optimal form of textual inputs (\S\ref{sec_optimal_form_of_textual_inputs}), more failure cases of our approach (\S\ref{sec_more_failure_cases}), some results on rectangular datasets (\S\ref{sec_results_on_rectangular_dataset}), and discussions about the societal impact (\S\ref{sec_societal_impact}). 

\section{Cross-Attention (CA) in Stable Diffusion}
\label{sec_CA}

\myparagraphsupp{This is supplementary to Section~\textcolor{red}{4.1} ``\textbf{rectifying diffusion model}''.} In this section, we provide an elaboration of Cross-Attention (CA) for a clearer comparison with our proposed Rectified Cross-Attention (RCA). For a CA layer in Stable Diffusion, let $\varphi_{I}$ and $\varphi_{T}$ denote the input image feature and text embeddings, respectively. Image queries $Q$, text keys $K$, and text values $V$ can be calculated by:
\begin{equation}
  Q=W_{Q}\cdot\varphi_{I}, \; K=W_{K}\cdot\varphi_{T}, \; V=W_{V}\cdot\varphi_{T},
  \label{supp_eq1}
\end{equation}
where $W_{Q}$, $W_{K}$, and $W_{V}$ are learnable projection matrices. Then attention score maps $\mathcal{M}$ are computed as:
\begin{equation}
  \mathcal{M}=\frac{QK^T}{\sqrt{d}}\in\mathbb{R}^{C\times H\times W},
  \label{supp_eq2}
\end{equation}
where $d$ is the scaling factor that is set as the dimension of the queries and keys, and $C$, $H$, $W$ are the channel number, height, and weight of $\mathcal{M}$, respectively. After that, we can calculate the output image feature $\mathcal{O}$ of this CA layer by:
\begin{equation}
  \mathcal{O}=\text{softmax}(\mathcal{M})V.
  \label{supp_eq3}
\end{equation}

\begin{figure}[t]
   \centering
   \includegraphics[width=1\linewidth]{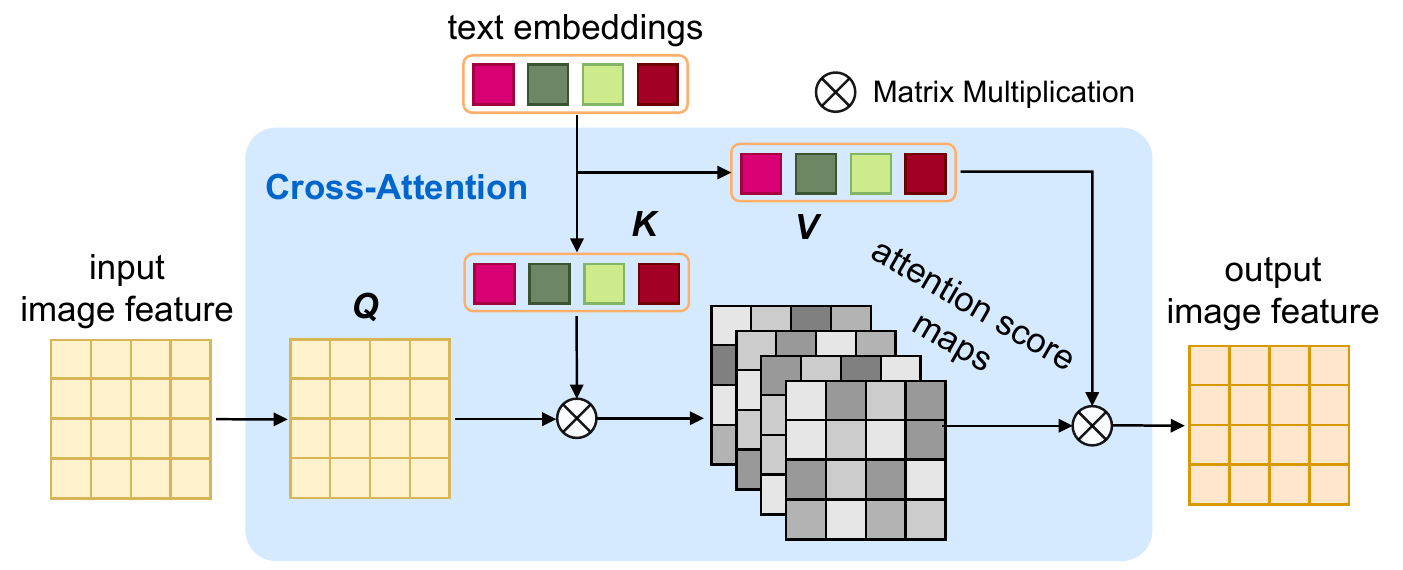}
   %\vspace{-0.1cm}
   \caption{An illustration of Cross-Attention (CA). 
   }
   \label{fig:CA}
%   \vspace{-0.4cm}
\end{figure}

A visual illustration of CA is shown in Figure~\ref{fig:CA}. In contrast, the proposed RCA rectifies $\mathcal{M}$ via \cref{eq2} in the main paper before applying softmax.

\section{Algorithm}
\label{sec_algorithm}

\myparagraphsupp{This is supplementary to Section~\textcolor{red}{4.1} ``\textbf{rectifying diffusion model}''.} The computation pipeline of RCA is illustrated in Algorithm~\ref{alg}.

\section{Additional implementation details}
\label{sec_additional_implementation_details}
\myparagraphsupp{This is supplementary to Section~\textcolor{red}{5.1} ``\textbf{experimental settings}''.} Training on COCO-Stuff/ADE20K takes about 6/2 days on a single NVIDIA A100 GPU. All our experiments are conducted using Stable Diffusion v1.4.

\begin{algorithm}
\caption{RCA}
\label{alg}
\SetAlgoLined
\SetKwInput{KwData}{Input}
\SetKwInput{KwResult}{Output}
 \KwData{Input image feature $\varphi_{I}$, text embeddings $\varphi_{T}$, and layout $l$}
 \KwResult{Output image feature $\mathcal{O}$}
Get image queries $Q$, text keys $K$, and text values $V$ by \cref{supp_eq1}\\
Get attention score maps $\mathcal{M}\in\mathbb{R}^{C\times H\times W}$ by \cref{eq1}\\
Initialize a mask $L\in\mathbb{R}^{C\times H\times W}$\\
Resize $l$ to match the spatial size of $\mathcal{M}$\\
\For{$k$ \emph{\textbf{in}} $\{0,1,...,C-1\}$}
{
\uIf{The $k$-th text embedding corresponds to a concept $m$}{
Find the binary map $l^{m}\in\mathbb{R}^{H\times W}$ in $l$ corresponding to this concept\\
$L^{k} \leftarrow l^{m}$\\
}
\Else{
$L^{k} \leftarrow 1$\\
}
}
Get the rectified attention score maps $\widehat{\mathcal{M}}$ by \cref{eq2}\\
Get the output image feature $\mathcal{O}$ by \cref{eq3}
\end{algorithm}

\section{More qualitative results on FLIS}
\label{sec_more_qualitative_results_on_FLIS}

\begin{figure*}[t]
   \centering
   \includegraphics[width=1\linewidth]{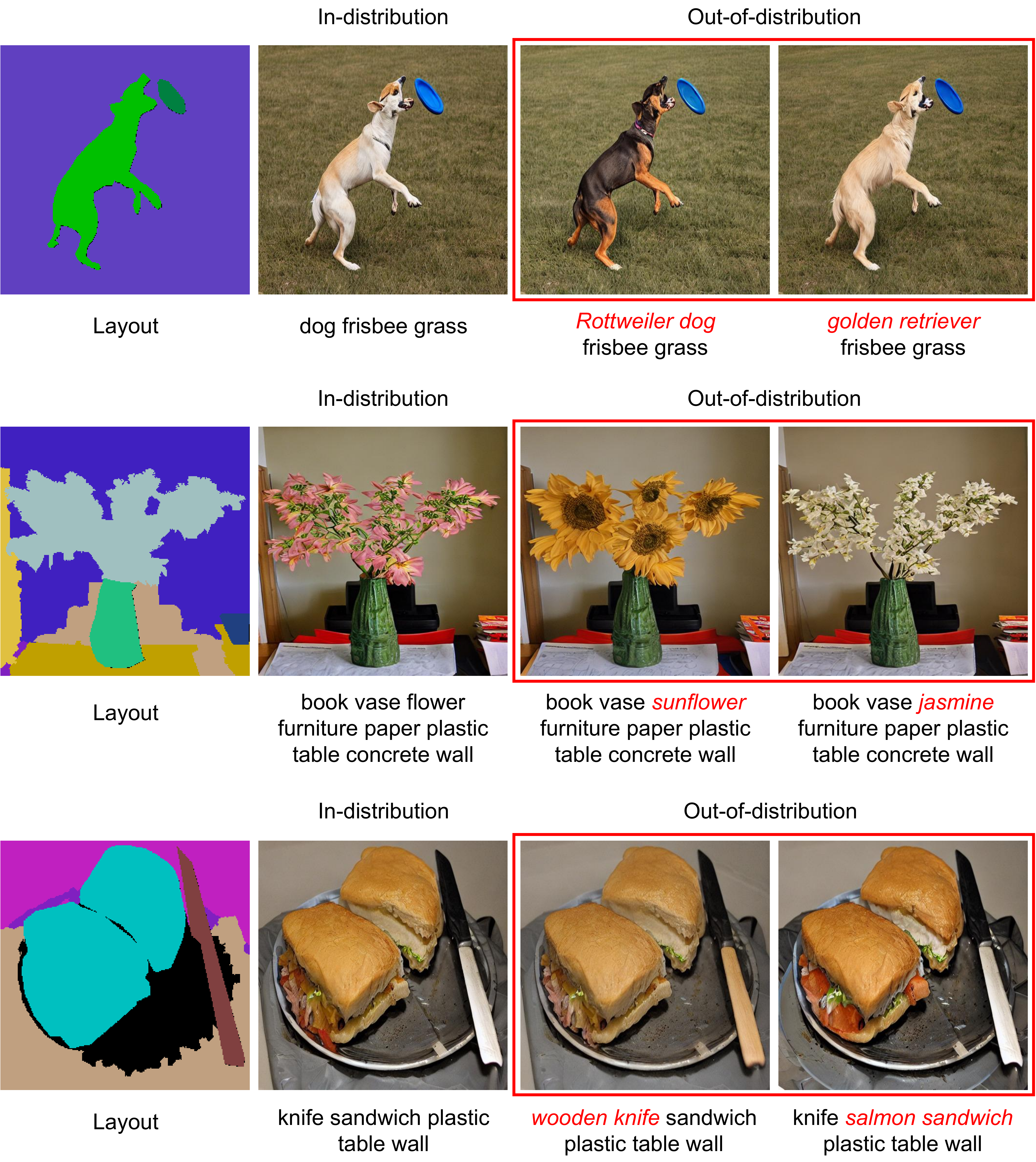}
   %\vspace{-0.1cm}
   \caption{\mycaptionsupp{Supplementary to Figure~\ref{fig:my_results}.} Our FreestyleNet is able to bind new attributes to the
    objects.
   }
   \label{fig:attribute}
%   \vspace{-0.4cm}
\end{figure*}

\begin{figure*}[t]
   \centering
   \includegraphics[width=1\linewidth]{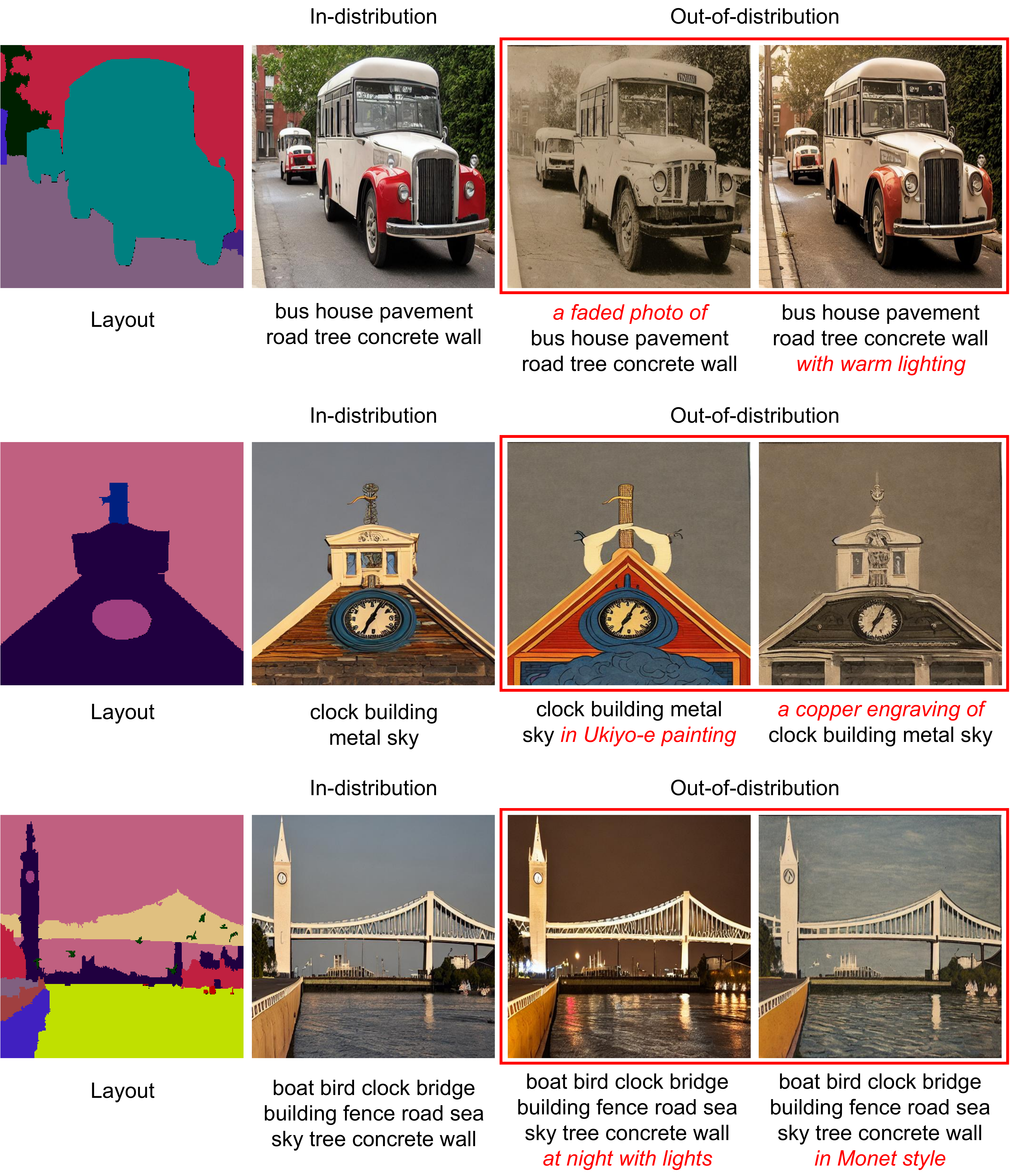}
   %\vspace{-0.1cm}
   \caption{\mycaptionsupp{Supplementary to Figure~\ref{fig:my_results}.} Our FreestyleNet is able to specify the styles for the synthesized images.
   }
   \label{fig:style}
%   \vspace{-0.4cm}
\end{figure*}

\begin{figure*}[t]
   \centering
   \includegraphics[width=1\linewidth]{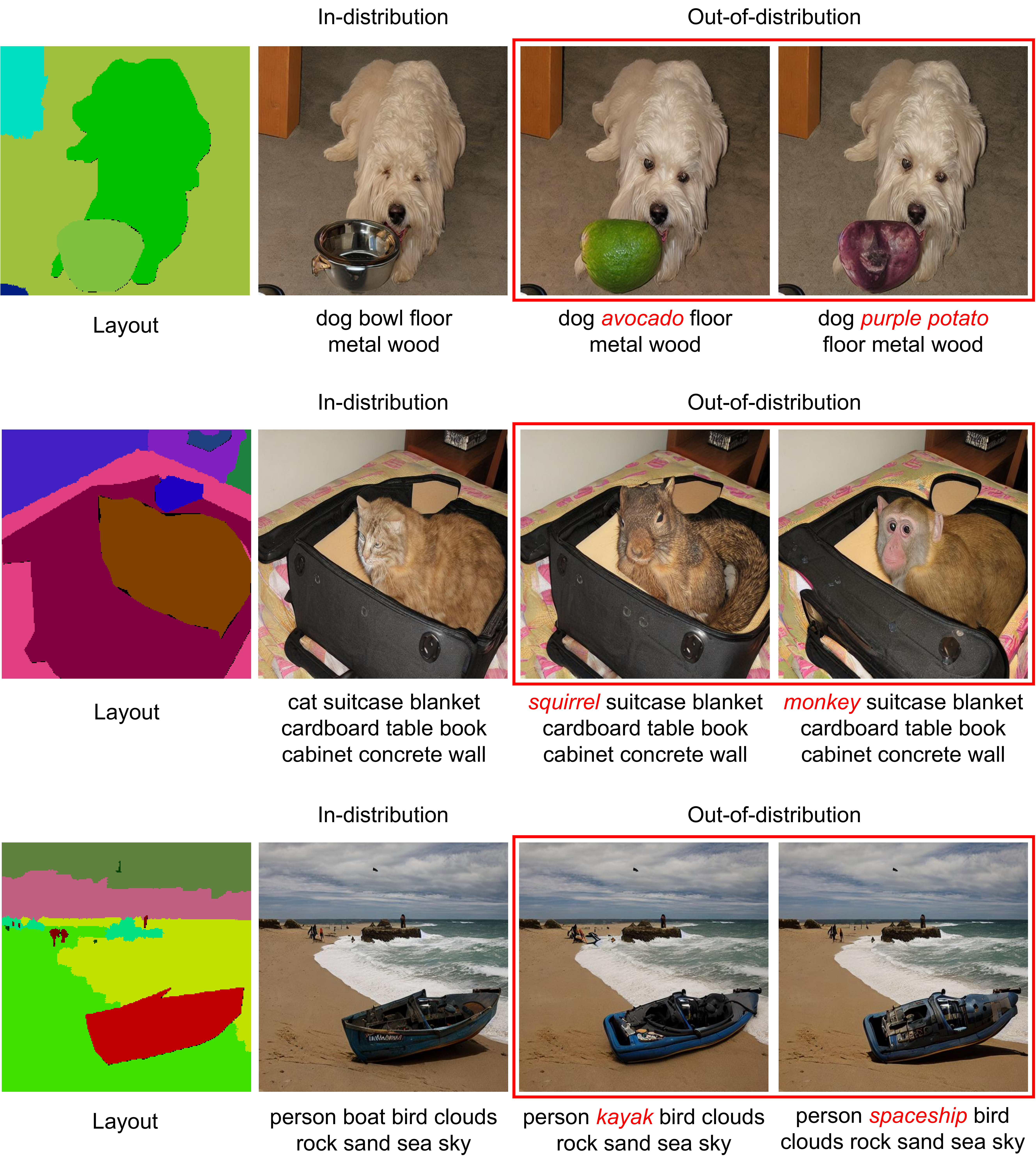}
   %\vspace{-0.1cm}
   \caption{\mycaptionsupp{Supplementary to Figure~\ref{fig:my_results}.} Our FreestyleNet is able to generate unseen objects.
   }
   \label{fig:object}
%   \vspace{-0.4cm}
\end{figure*}

\myparagraphsupp{This is supplementary to Section~\textcolor{red}{5.2} ``\textbf{qualitative evaluation on FLIS}''.}
In Figures~\ref{fig:attribute}, \ref{fig:style}, and \ref{fig:object}, we present more FLIS results by using the proposed model. They demonstrate the capability of our method for FLIS and its high potential to spawn various applications.

\section{More comparisons with LIS baselines} 
\label{sec_more_comparisons_with_LIS_baselines}

\begin{figure*}[t]
   \centering
   \includegraphics[width=0.96\linewidth]{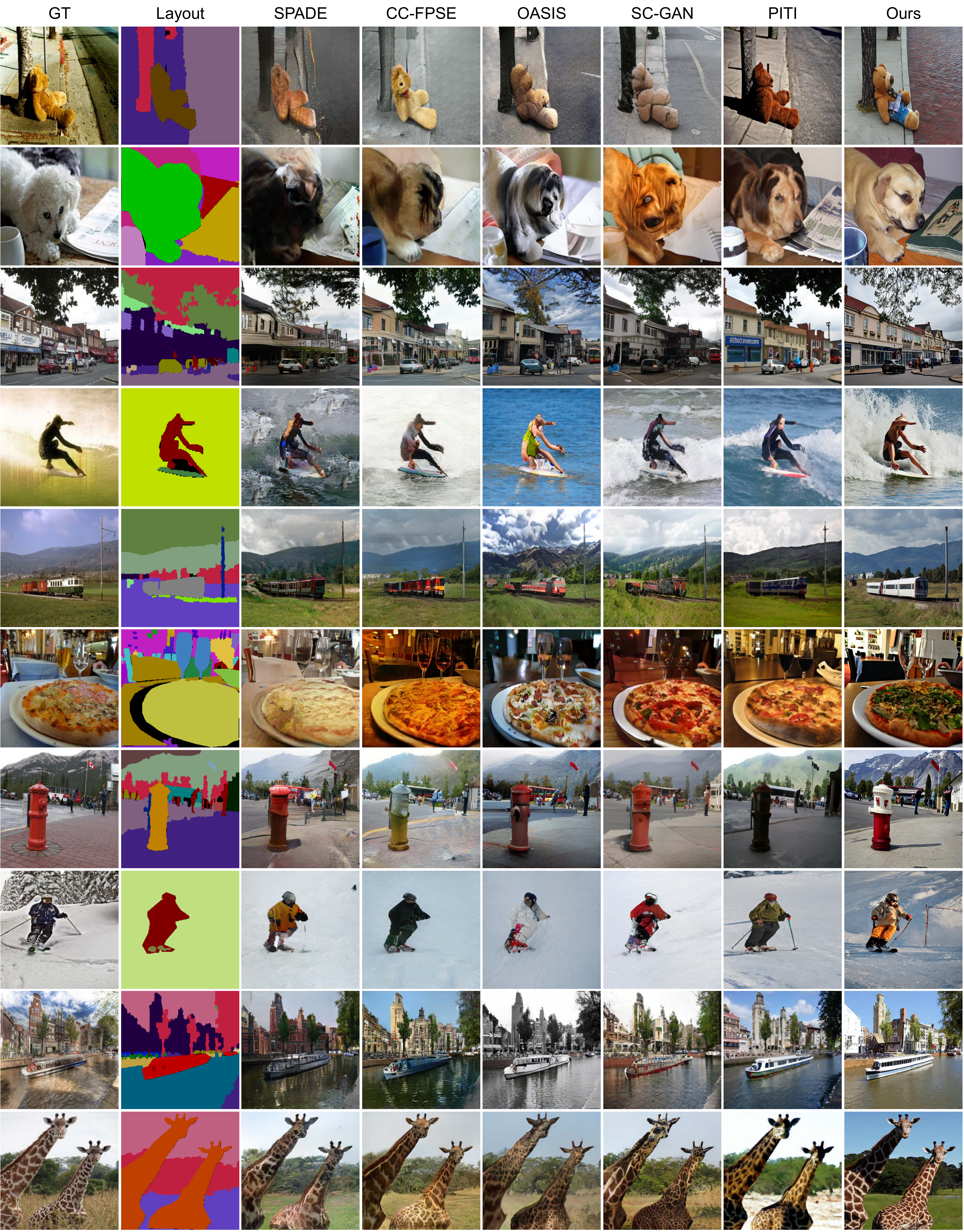}
   %\vspace{-0.1cm}
   \caption{\mycaptionsupp{Supplementary to Figure~\ref{fig:comparison}.} Visual comparison results with LIS baselines on COCO-Stuff.
   }
   \label{fig:comp_coco}
%   \vspace{-0.4cm}
\end{figure*}

\begin{figure*}[t]
   \centering
   \includegraphics[width=0.96\linewidth]{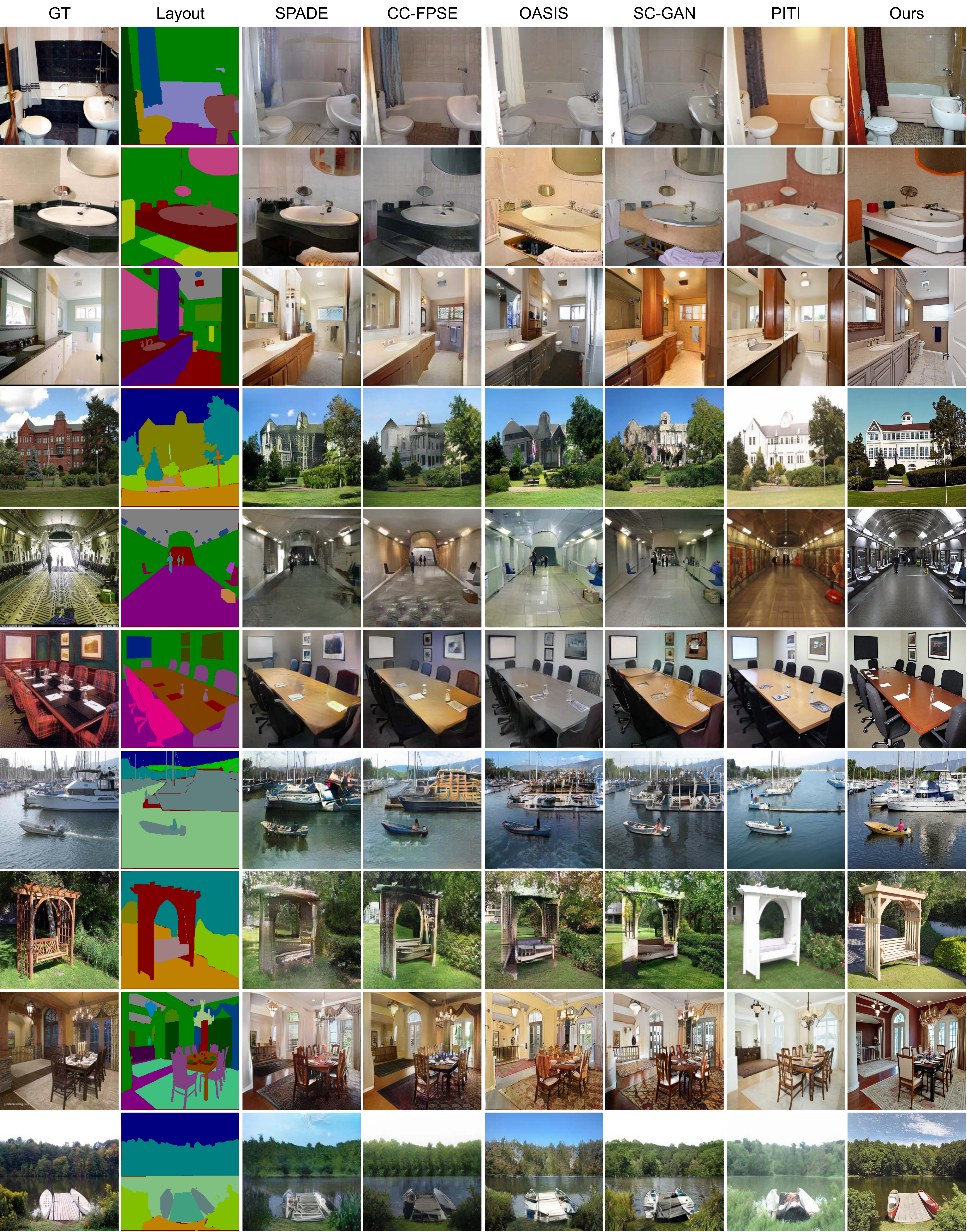}
   %\vspace{-0.1cm}
   \caption{\mycaptionsupp{Supplementary to Figure~\ref{fig:comparison}.} Visual comparison results with LIS baselines on ADE20K.
   }
   \label{fig:comp_ade20k}
%   \vspace{-0.4cm}
\end{figure*}

\myparagraphsupp{This is supplementary to Section~\textcolor{red}{5.3} ``\textbf{comparison with LIS baselines}''.} 
In this section, we provide more comparison results between SPADE~\cite{park2019semantic}, CC-FPSE~\cite{liu2019learning}, OASIS~\cite{sushko2020you}, SC-GAN~\cite{wang2021image}, PITI~\cite{wang2022pretraining}, and our method. Figures~\ref{fig:comp_coco} and \ref{fig:comp_ade20k} show the results on COCO-Stuff~\cite{caesar2018coco} and ADE20K~\cite{zhou2017scene}, respectively. These results indicate the superiority of our method in generating high-fidelity images in the context of LIS.

\begin{figure*}[t]
   \centering
   \includegraphics[width=0.9\linewidth]{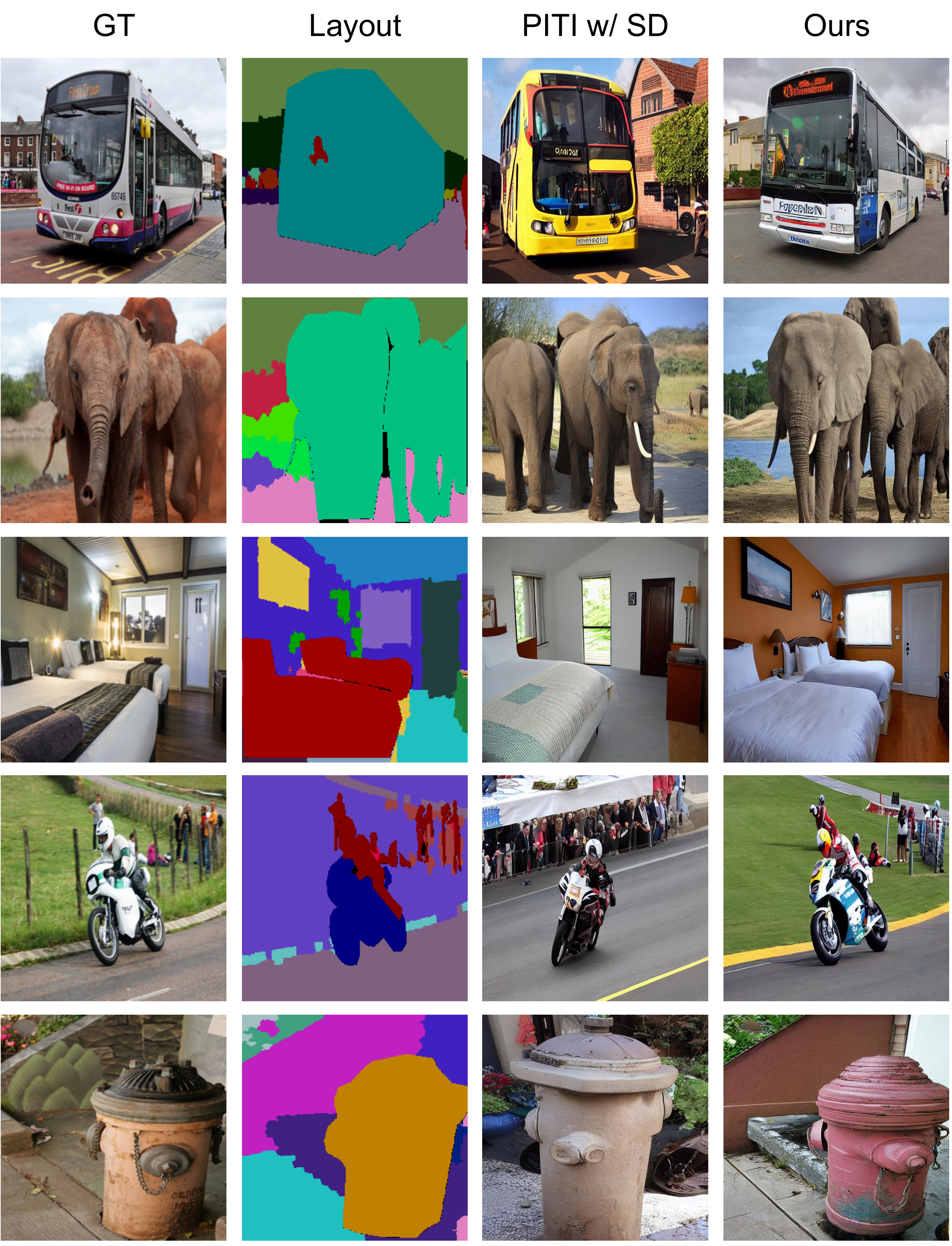}
   %\vspace{-0.1cm}
   \caption{PITI w/ SD represents the PITI method whose diffusion model (GLIDE) is replaced by Stable Diffusion.
   }
   \label{fig:piti-wsd}
%   \vspace{-0.4cm}
\end{figure*}

For a fair comparison with PITI, we replace its pre-trained text-to-image diffusion model (GLIDE~\cite{nichol2021glide}) with Stable Diffusion~\cite{rombach2022high}. Due to time limits, we carefully tune learning rates only when training its model (we call it PITI w/ SD). Some visual results are provided in Figure~\ref{fig:piti-wsd}. The images synthesized by PITI w/ SD exhibit good visual quality but the spatial alignment with the input layout is poor (clearly poorer than ours). The quantitative comparison results are also provided in Table~\ref{tab:fid_comp}.

Here we compare our FreestyleNet with additional related works including Lab2Pix-V2~\cite{zhu2022label}, sVQGAN-T~\cite{alaniz2022semantic}, and PoE-GAN~\cite{huang2022multimodal}. The comparison results under the in-distribution setting is reported in Table~\ref{tab:addi_comp}. As neither sVQGAN-T~\cite{alaniz2022semantic} nor PoE-GAN~\cite{huang2022multimodal} provide code, their results are copied from their papers. These results showcase our superiority over the others.

\section{Diversity evaluation} 
\label{sec_diversity_evaluation}

\begin{figure*}[t]
   \centering
   \includegraphics[width=1\linewidth]{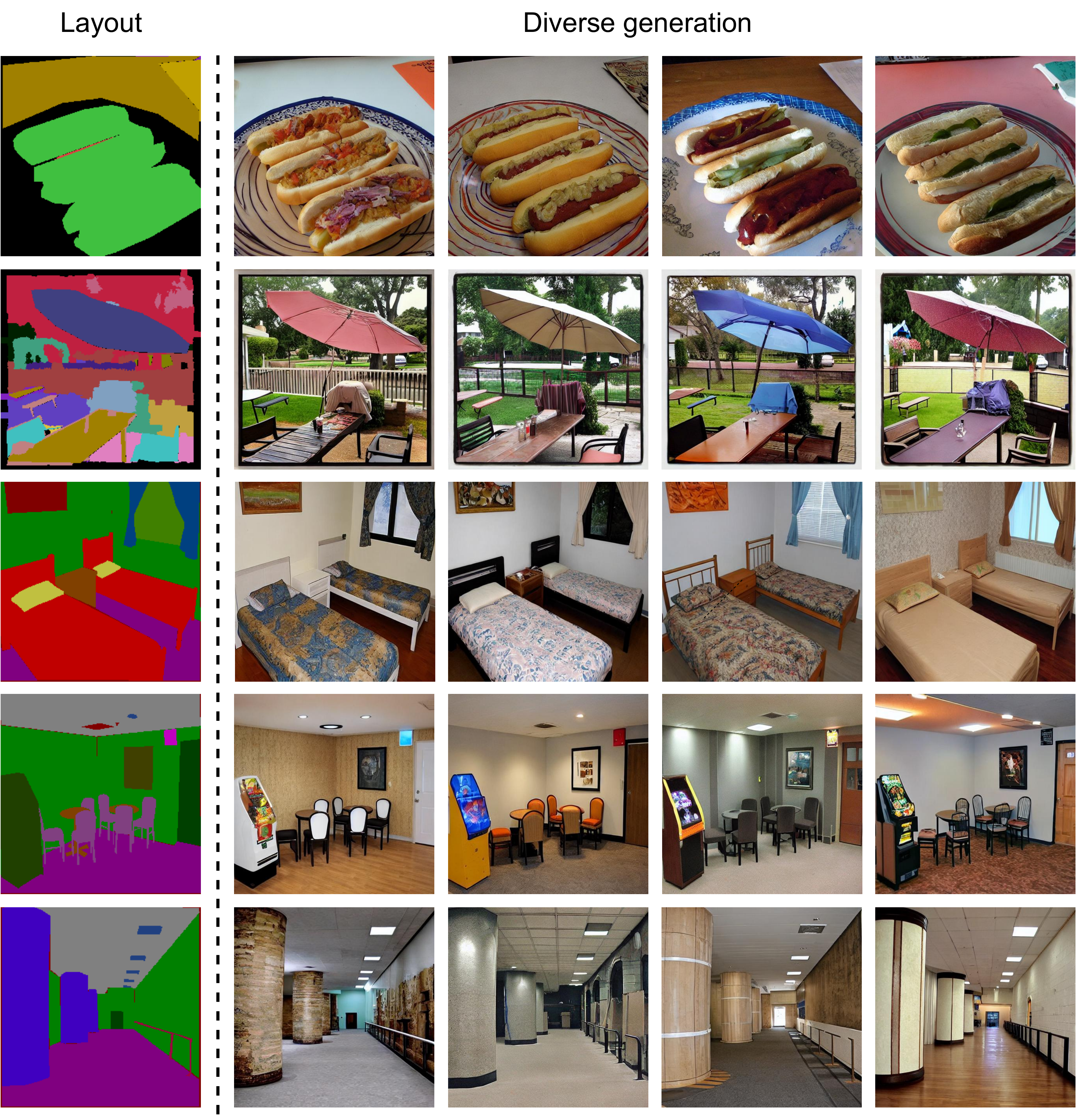}
   %\vspace{-0.1cm}
   \caption{Diverse generation results of our FreestyleNet. 
   }
   \label{fig:diversity}
%   \vspace{-0.4cm}
\end{figure*}

\myparagraphsupp{This is supplementary to Section~\textcolor{red}{5.3} ``\textbf{comparison with LIS baselines}''.} In this section, we conduct some experiments to evaluate the generation diversity of different methods. Note that our model naturally enables generation with high diversity from the same layout by using various texts (see Figures~\ref{fig:teaser}, \ref{fig:my_results}, and \ref{fig:effec_RCA} in the main paper). Here we perform the diversity evaluation in the conventional LIS setting. Following OASIS~\cite{sushko2020you}, we calculate LPIPS~\cite{zhang2018unreasonable} between images generated from the same layout (and same text for our model) but with randomly sampled noise. The evaluation results are provided in Table~\ref{tab:diver_eval}. Our model achieves the highest LPIPS among all comparison methods. We also show some visual samples in Figure~\ref{fig:diversity}.

\begin{table}
\setlength{\tabcolsep}{5pt}
  \centering
  \begin{tabular}{ccc}
    \toprule[0.8pt]
    Method & PITI w/ SD & FreestyleNet (ours) \\
    \midrule[0.5pt]
    FID$\downarrow$ & 15.5 & \textbf{14.4} \\
    mIoU$\uparrow$ & 13.1 & \textbf{40.7} \\
    \bottomrule[0.8pt]
  \end{tabular}
  %\vspace{2mm}
  \caption{Quantitative comparison results with PITI w/ SD on COCO-Stuff.}
  \label{tab:fid_comp}
%   \vspace{-0.4cm}
\end{table}
\begin{table}
\setlength{\tabcolsep}{4.5pt}
  \centering
  \begin{tabular}{cccccc}
    \toprule[0.8pt]
    \multirow{2.5}*{Method} & \multicolumn{2}{c}{COCO-Stuff} & \textbf{} & \multicolumn{2}{c}{ADE20K} \\
    \cmidrule[0.5pt]{2-3}
    \cmidrule[0.5pt]{5-6}
    ~ & FID$\downarrow$ & mIoU$\uparrow$ & \textbf{} & FID$\downarrow$ & mIoU$\uparrow$ \\
    \midrule[0.5pt]
    Lab2Pix-V2~\cite{zhu2022label} & 18.1 & 40.5 & \textbf{} & 31.3 & 41.0 \\
    sVQGAN-T~\cite{alaniz2022semantic} & 28.8 & - & \textbf{} & 38.4 & - \\
    PoE-GAN~\cite{huang2022multimodal} & 15.8 & - & \textbf{} & - & - \\
    FreestyleNet (ours) & \textbf{14.4} & \textbf{40.7} & \textbf{} & \textbf{25.0} & \textbf{41.9} \\
    \bottomrule[0.8pt]
  \end{tabular}
  %\vspace{2mm}
  \caption{Comparison results with additional related works.}
  \label{tab:addi_comp}
%   \vspace{-0.4cm}
\end{table}

\section{Optimal form of textual inputs}
\label{sec_optimal_form_of_textual_inputs}
\myparagraphsupp{This is supplementary to Section~\textcolor{red}{4.1} ``\textbf{rectifying diffusion model}''.} As full-form image descriptions are expensive (or even intractable) to collect, we suggest using the stacked concepts which can be easily obtained from semantic labels. Moreover, stacked concepts fit naturally into the design of RCA, which builds the relationship between each individual semantic and its position on the image. We actually have explored several alternatives (which perform worse), including (1) keyword-to-sentence translation, (2) learnable prompts, and (3) manual construction of full-form prompts for inference. We believe that looking for the optimal form of textual inputs is important, and we will explore it for future work.
\begin{table}
\setlength{\tabcolsep}{5.6pt}
  \centering
  \begin{tabular}{ccc}
    \toprule[0.8pt]
    \multirow{2.5}*{Method} & \multicolumn{2}{c}{LPIPS$\uparrow$} \\
    \cmidrule[0.5pt]{2-3}
    ~ & COCO-Stuff & ADE20K \\
    \midrule[0.5pt]
    CC-FPSE~\cite{liu2019learning} & 0.089 & 0.129 \\
    OASIS~\cite{sushko2020you} & 0.345 & 0.285\\
    PITI~\cite{wang2022pretraining} & 0.523 & 0.480\\
    FreestyleNet (ours) & \textbf{0.592} & \textbf{0.591} \\
    \bottomrule[0.8pt]
  \end{tabular}
  %\vspace{2mm}
  \caption{Diversity evaluation results. Pix2PixHD~\cite{wang2018high}, SPADE~\cite{park2019semantic}, and SC-GAN~\cite{wang2021image} do not support diverse generation (\ie, LPIPS is 0).}
  \label{tab:diver_eval}
%   \vspace{-0.4cm}
\end{table}

\section{More failure cases} 
\label{sec_more_failure_cases}

\begin{figure*}[t]
   \centering
   \includegraphics[width=0.66\linewidth]{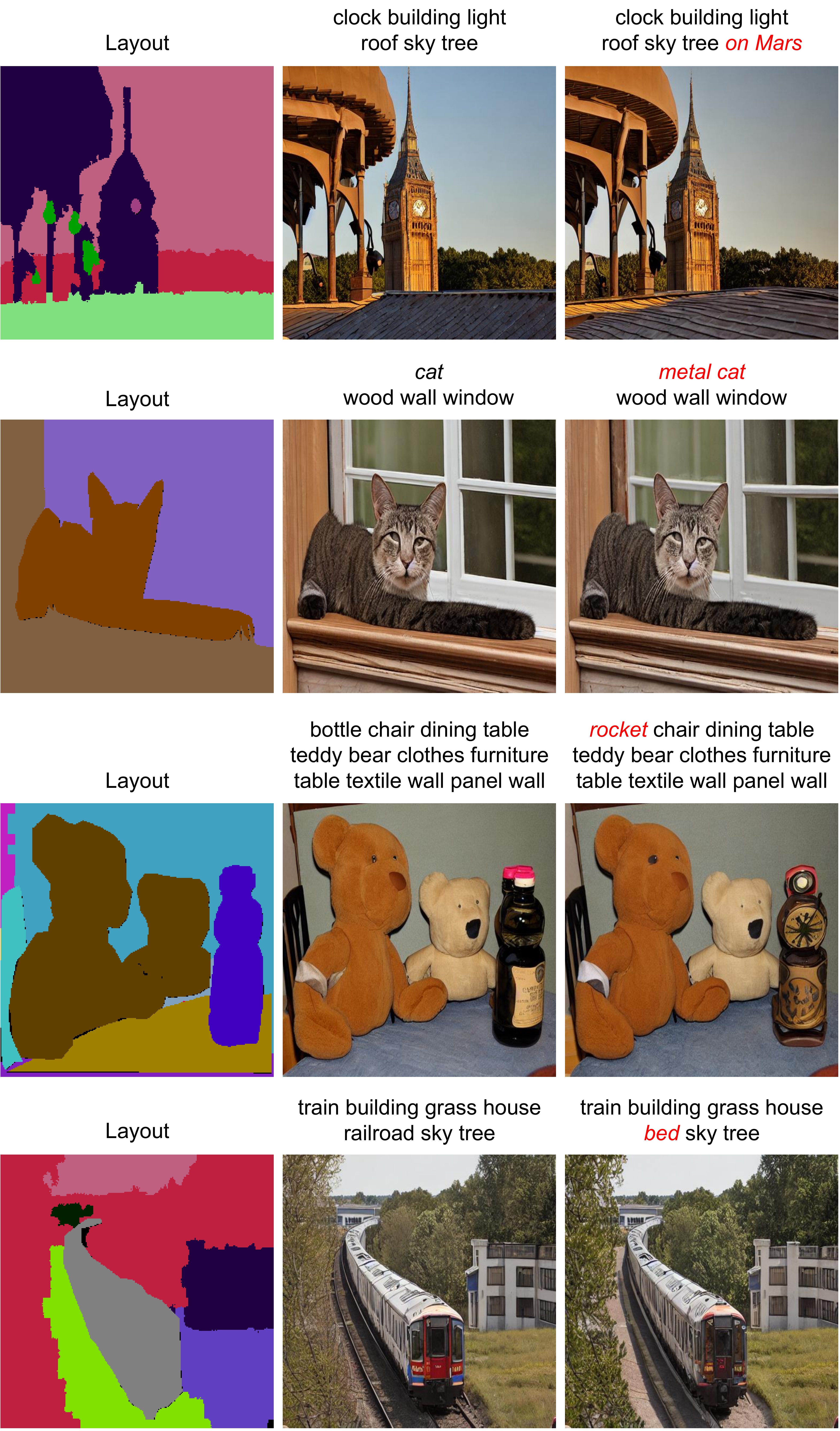}
   %\vspace{-0.1cm}
   \caption{\mycaptionsupp{Supplementary to Figure~\ref{fig:failure_case}.} Failure cases. It is difficult for our FreestyleNet to generate some rare semantics or unreasonable scenes.
   }
   \label{fig:failure}
%   \vspace{-0.4cm}
\end{figure*}

\myparagraphsupp{This is supplementary to Section~\textcolor{red}{5.5} ``\textbf{limitations}''.} 
In Figure~\ref{fig:failure}, we show more failure cases of the proposed model. These results are in line with our conclusion that our method sometimes fails to synthesize counterfactual scenes. This limitation can possibly be alleviated in our future work, by 1) leveraging more powerful pre-trained text-to-image models, and 2) investigating better ways to retain the generative capability of the pre-trained model, perhaps by prompting techniques.

\section{Results on rectangular datasets} 
\label{sec_results_on_rectangular_dataset}
\begin{figure*}[t]
   \centering
   \includegraphics[width=0.82\linewidth]{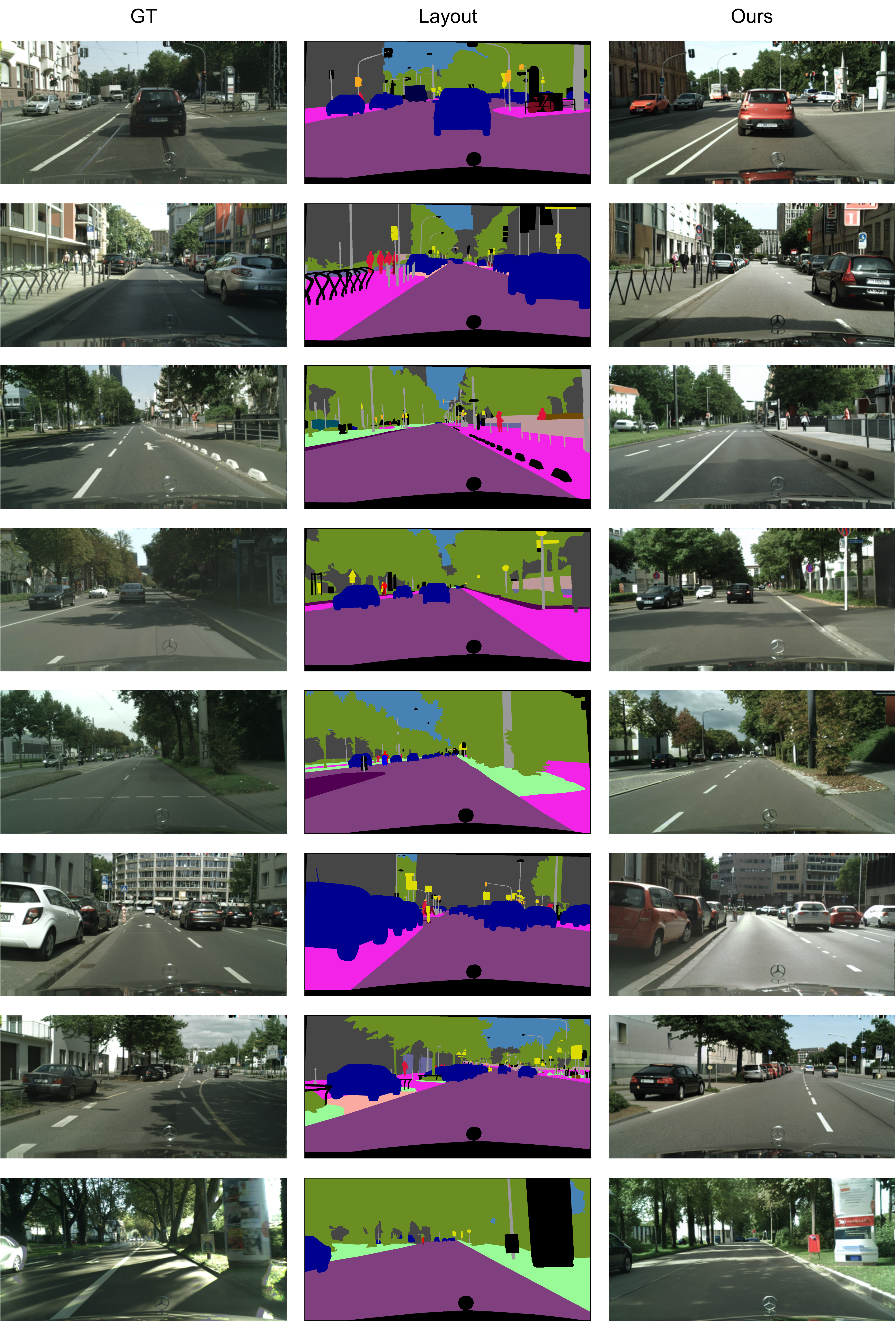}
   %\vspace{-0.1cm}
   \caption{Generation results of our FreestyleNet on Cityscapes. 
   }
   \label{fig:cityscapes}
%   \vspace{-0.4cm}
\end{figure*}

The pre-trained Stable Diffusion that we leverage is designed to generate square (512$\times$512) images. To verify the validity of the proposed method on rectangular datasets, we train our model on Cityscapes~\cite{cordts2016cityscapes}. We resize all images of Cityscapes to 512$\times$512 during training and resize the synthesized results back to the original size in testing phase. As shown in Figure~\ref{fig:cityscapes}, our method yields visually pleasing results.

\section{Societal impact} 
\label{sec_societal_impact}
Our method allows the users to generate diverse images using text and layout. This ability may be maliciously used for  content, which incurs potential negative social impacts such as the spread of fake news and invasion of privacy. To mitigate them, powerful deepfake detection methods that automatically distinguish deepfake images from real ones are needed.

% \clearpage
% \clearpage
% \clearpage


\end{document}